\documentclass[sigconf, nonacm]{acmart}

\newcommand\vldbdoi{10.14778/3476249.3476270}
\newcommand\vldbpages{XXX-XXX}
\newcommand\vldbvolume{14}
\newcommand\vldbissue{11}
\newcommand\vldbyear{2021}

\newcommand\vldbtitle{\shorttitle} 
\newcommand\vldbavailabilityurl{http://vldb.org/pvldb/format_vol14.html}

\usepackage{xspace}
\newcommand{\sys}{\textsc{VolcanoML}\xspace}

\usepackage{subfigure}
\usepackage{multirow}
\usepackage{bm}
\usepackage{bbm}
\usepackage{color, soul}
\usepackage{listings}
\usepackage{xcolor}
\usepackage{multibib}
\newcites{A}{References}

\usepackage[linesnumbered,vlined,ruled,noend]{algorithm2e}

\newcommand{\para}[1]{{\vspace{2pt} \bf \noindent #1 \hspace{1pt}}}
\definecolor{codegray}{rgb}{0.5,0.5,0.5}

\usepackage{enumitem}

\usepackage{wrapfig}

\begin{document}
\title{VolcanoML: Speeding up End-to-End AutoML via 
\\Scalable Search Space Decomposition}
\subtitle{[Scalable Data Science]}

\renewcommand{\shorttitle}{VolcanoML: Speeding up End-to-End AutoML via Scalable Search Space Decomposition}


\author{Yang Li$^{\dagger\ddagger}$, Yu Shen$^\dagger$, Wentao Zhang$^\dagger$, Jiawei Jiang$^\ddagger$, Bolin Ding$^\mathsection$, Yaliang Li$^\mathsection$}
\author{Jingren Zhou$^\mathsection$, Zhi Yang$^\dagger$, Wentao Wu$^*$, Ce Zhang$^\ddagger$, Bin Cui$^\dagger$}
\affiliation{
$^\dagger$EECS, Peking University~~~~~$^\ddagger$ETH Zurich~~~~~$^\mathsection$Alibaba Group~~~~~$^*$Microsoft Research, Redmond
}
\affiliation{
$^\dagger$\{liyang.cs, shenyu, wentao.zhang, yangzhi, bin.cui\}@pku.edu.cn $^\ddagger$\{jiawei.jiang, ce.zhang\}@inf.ethz.ch~~~~~\\$^\mathsection$\{bolin.ding, yaliang.li, jingren.zhou\}@alibaba-inc.com~~~~~$^*$wentao.wu@microsoft.com
}

\renewcommand{\shortauthors}{Li et al.}

\begin{abstract}
End-to-end AutoML has attracted intensive interests from both academia and industry, which automatically searches for ML pipelines in a space induced by feature engineering, algorithm/model selection, and hyper-parameter tuning.
Existing AutoML systems, however, suffer from scalability issues when applying to application domains with large, high-dimensional search spaces.
We present \sys, a scalable and extensible framework that facilitates systematic exploration of large AutoML search spaces.
\sys introduces and implements basic building blocks that decompose a large search space into smaller ones, and allows users to utilize these building blocks to compose an \emph{execution plan} for the AutoML problem at hand.
\sys further supports a Volcano-style \emph{execution model} -- akin to the one supported by modern database systems -- to execute the plan constructed.
Our evaluation demonstrates that, not only does \sys raise the level of expressiveness for search space decomposition in AutoML, it also leads to actual findings of decomposition strategies that are significantly more efficient than the ones employed by state-of-the-art AutoML systems such as \texttt{auto-sklearn}.
\end{abstract}

\settopmatter{printfolios=true}
\maketitle

\begingroup\small\noindent\raggedright\textbf{PVLDB Reference Format:}\\
Yang Li, Yu Shen, Wentao Zhang, Jiawei Jiang, Bolin Ding, Yaliang Li, Jingren Zhou, Zhi Yang, Wentao Wu, Ce Zhang, and Bin Cui. \vldbtitle. PVLDB, \vldbvolume(\vldbissue): \vldbpages, \vldbyear.\\
\href{https://doi.org/\vldbdoi}{doi:\vldbdoi}
\endgroup
\begingroup
\renewcommand\thefootnote{}\footnote{\noindent
This work is licensed under the Creative Commons BY-NC-ND 4.0 International License. Visit \url{https://creativecommons.org/licenses/by-nc-nd/4.0/} to view a copy of this license. For any use beyond those covered by this license, obtain permission by emailing \href{mailto:info@vldb.org}{info@vldb.org}. Copyright is held by the owner/author(s). Publication rights licensed to the VLDB Endowment. \\
\raggedright Proceedings of the VLDB Endowment, Vol. \vldbvolume, No. \vldbissue\ %
ISSN 2150-8097. \\
\href{https://doi.org/\vldbdoi}{doi:\vldbdoi} \\
}\addtocounter{footnote}{-1}\endgroup

\ifdefempty{\vldbavailabilityurl}{}{
\vspace{.3cm}
\begingroup\small\noindent\raggedright\textbf{PVLDB Availability Tag:}\\
The source code of this research paper has been made publicly available at \textbf{\url{https://github.com/PKU-DAIR/mindware}}.
\endgroup
}

\section{Introduction}
\label{sec:intro}


In recent years, researchers in the database community have been working on raising the level of abstractions of machine learning (ML) and integrating such functionality into today's data management systems,
e.g., SystemML~\cite{Ghoting2011}, SystemDS~\cite{Boehm2019}, Snorkel~\cite{snorkel}, ZeroER~\cite{ZeroER}, TFX~\cite{baylor2017tfx, TFX}, ``Query 2.0''~\cite{Query20}, Krypton~\cite{Krypton}, Cerebro~\cite{Cerebro}, ModelDB~\cite{modeldb}, MLFlow~\cite{mlflow}, DeepDive~\cite{DeepDive}, HoloClean~\cite{Holoclean}, ActiveClean~\cite{ActiveClean}, and NorthStar~\cite{NorthStar}.
End-to-end AutoML systems~\cite{automl,DBLP:journals/corr/abs-1904-12054,automl_book} have been an emerging type of systems that has significantly raised the level of abstractions of building ML applications. 
Given an input dataset and a user-defined utility metric (e.g., validation accuracy), these systems automate the search of an end-to-end ML pipeline, including \textit{feature engineering}, \textit{algorithm/model selection}, and \textit{hyper-parameter tuning}. 
Open-source examples include \texttt{auto-sklearn}~\cite{feurer2015efficient}, \texttt{TPOT}~\cite{olson2019tpot}, and \texttt{hyperopt-sklearn}~\cite{komer2014hyperopt}, whereas most cloud service providers, e.g., Google, Microsoft, Amazon, Alibaba, etc., all provide their proprietary services on the cloud.
As machine learning has become an increasingly indispensable functionality integrated in modern data (management) systems, an efficient and effective end-to-end AutoML component also becomes increasingly important.

End-to-end AutoML provides a powerful abstraction to automatically navigate and search in a given complex search space. 
However, in our experience of applying state-of-the-art end-to-end AutoML systems in a range of real-world applications, we find that such a system running ``fully automatically'' is rarely enough --- often, developing a successful ML application involves multiple iterations between a user and an AutoML system to iteratively improve the resulting ML artifact.

\vspace{-0.5em}
\noindent
\paragraph*{\bf Motivating Practical Challenge}
One such type of interaction, which inspires this work, is the \textit{\underline{enrichment of search space}}.
We observe that the default search space provided by state-of-the-art AutoML systems is often not enough in 
many applications. 
This was not obvious to us at all in the beginning and it is not until we finish building a range of real-world applications that we realize this via a set of concrete examples. 
For example, in one of our astronomy applications~\cite{app1}, the feature normalization function is domain-specific and not supported by most, if not all, AutoML systems. 
Similar examples can also be found when searching for suitable ML models via AutoML.
In one of our meteorology applications, we need to extend the models with meteorology-specific loss functions.
We saw similar problems when we tried to extend existing AutoML systems with pre-trained feature embeddings coming from TensorFlow Hub, to include newly arXiv'ed models to enrich the ``Model Base''~\cite{automl2}, or to support Cosine annealing as for tuning.

\vspace{-0.5em}
\paragraph*{\bf Technical Challenge: Scalability over the Search Space} 
``\textit{Why is it hard to extend the search space, as a user, in an end-to-end AutoML system}?'' The answer to this question is a complex one that is \textit{not} completely technical: 
some aspects are less technical such as engineering decisions and UX designs, however, there are also more \textit{technically fundamental} aspects. 
An end-to-end AutoML system contains an optimization algorithm that navigates a joint search space induced by \textit{feature engineering}, \textit{algorithm selection}, and \textit{hyper-parameter tuning}.
Because of this joint nature, the search space of end-to-end AutoML is complex and huge while the enrichment is only going to make it even larger.
As we will see, handling such a huge space is already challenging for existing systems, and further enriching it will make it even harder to scale.

\setlength{\columnsep}{0.5em}%
\begin{wrapfigure}{r}{0.2\textwidth}
  \begin{center}
    \vspace{-0.5em}
    \includegraphics[width=0.2\textwidth]{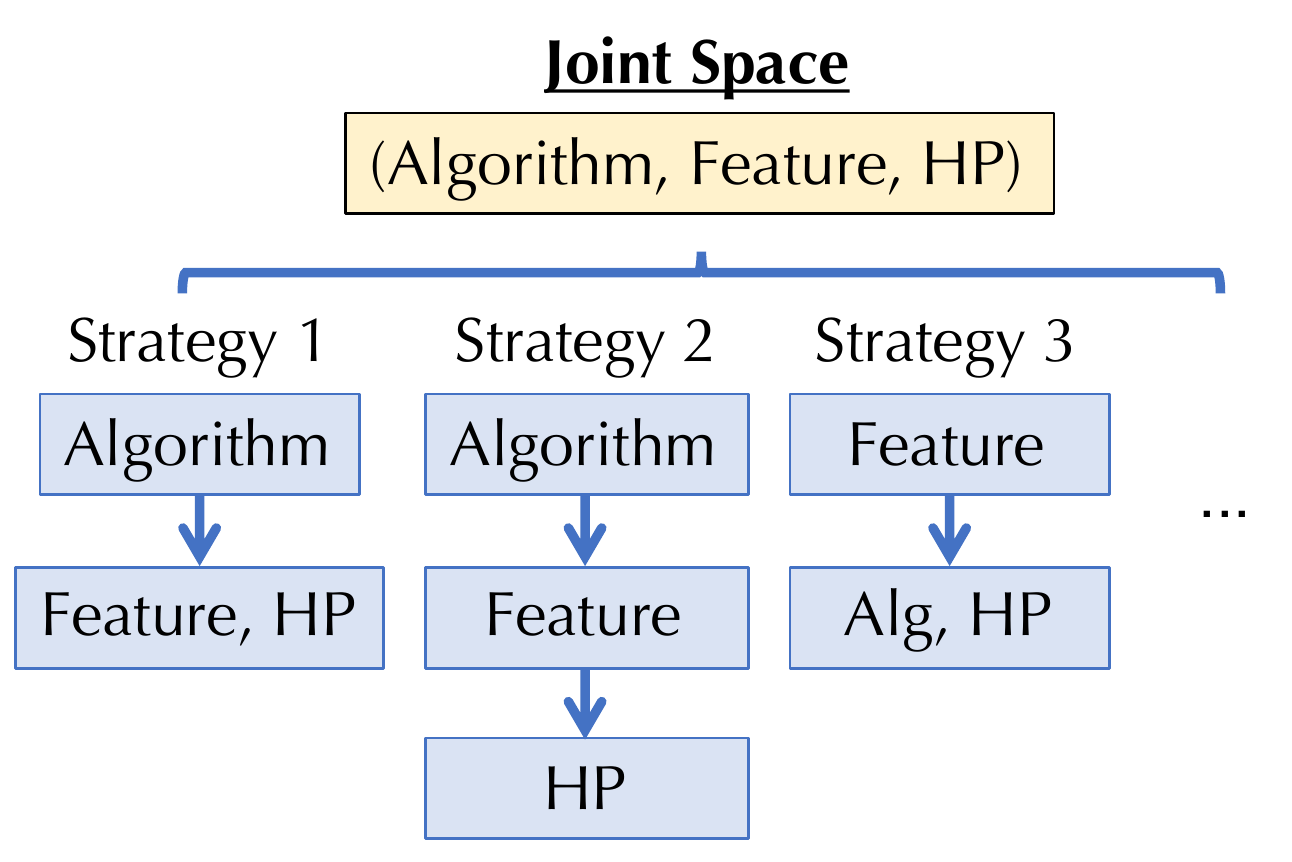}
  \end{center}
  \vspace{-1.5em}
\end{wrapfigure}

Many existing systems such as \texttt{auto-sklearn}~\cite{feurer2015efficient} and \texttt{TPOT}~\cite{olson2019tpot} deal with the entire composite search space \textit{jointly}, which naturally leads to the scalability bottleneck.
Decomposing a joint space has been explored for some subspaces (e.g., only algorithm and hyper-parameters as in ~\cite{liu2019automated, li2020efficient}), however, none of them has been applied to a search space as large as that of end-to-end AutoML. 
One challenge is that there exist many different ways to decompose the same space, as shown above, but only \textit{some} of them can perform well. 
\textit{Without a structured, high-level abstraction for search space decomposition to explore different strategies, it is very hard to scale up an end-to-end AutoML system to accommodate the search space that will only get larger in the future.}

\vspace{-0.5em}
\paragraph*{\bf Summary of Technical Contributions}
In this paper, we focus on designing a system, \sys, which is \textit{scalable to a large search space}. Our technical contributions are as follows.

\vspace{-0.5em}
\paragraph*{C1. System Design: A Structured View on Decomposition.} 
The main technical contribution of \sys is to provide a flexible and principled way of decomposing a large search space into multiple smaller ones.
We propose a novel system abstraction: a set of \textit{\sys building blocks} (Section~\ref{sec:building-blocks}), each of which takes charge of a smaller sub-search space whereas a \textit{\sys execution plan} (Section~\ref{sec:exec-plan}) consists of a \textit{tree} of such building blocks --- the root node corresponds to the original search space and its child nodes correspond to different subspaces. 
Under this abstraction, optimizing in the joint space is conducted as optimization problems over different smaller subspaces. 
The execution model is similar to the classic ``Volcano'' query evaluation model in a relational database~\cite{10.5555/1450931} (thus the name \sys): The system asks the root node to take one iteration in the optimization process, which \emph{recursively} invokes one of its child nodes to take one iteration on solving a smaller-scale optimization problem over its own subspace; this recursive invocation procedure will continue until a leaf node is reached.
This flexible abstraction allows us to explore different ways that the same joint space can be decomposed. Together with a range of additional optimizations (Section~\ref{sec:additional_optimization}), \sys can often support  more scalable search process than the existing AutoML systems such as \texttt{auto-sklearn} and \texttt{TPOT}.

\vspace{-0.5em}
\paragraph*{C2. Large-scale
Empirical Evaluations}

We conducted intensive empirical evalutions, comparing \sys with state-of-the-art systems including \texttt{auto-sklearn} and \texttt{TPOT}. 
We show that (1) under the \underline{\textit{same search space}} as \texttt{auto-sklearn}, \sys significantly outperforms \texttt{auto-sklearn} and \texttt{TPOT} --- over 30 classification tasks and 20 regression tasks --- \sys outperforms the \textit{best} of \texttt{auto-sklearn} and \texttt{TPOT} on a majority of tasks;
(2) using an \underline{\textit{enriched search space}} with additional feature engineering operators, \sys performs significantly better than \texttt{auto-sklearn};
and (3) using an \underline{\textit{enriched search space}} with an additional data processing stage and functionalities beyond what \texttt{auto-sklearn} and \texttt{TPOT} currently support (i.e., an additional embedding selection stage using pre-trained models on TensorFlow Hub), \sys can deal with input types such as images efficiently.

\vspace{-0.5em}
\paragraph*{\bf Moving Forward} The \sys abstraction enables a structured view of optimizing a black-box function via decomposition. This structured view itself opens up interesting future directions. 
For example, one may wish to \textit{automatically} decompose a search space given a workload, just like what a classic query optimizer would do for relational queries.
For constrained optimizations, we also imagine techniques similar to traditional ``\textit{push-down selection}'' could be applied in a similar spirit. 
We explore the possibility of automatically searching for the best plan in Section~\ref{sec:exec-plan} and discuss the limitations of this simple strategy and the exciting line of future work 
that could follow.
While the full treatment of these aspects are beyond the scope of this paper, we hope the \sys abstraction can serve as a foundation for these future endeavors.

\vspace{-0.5em}
\section{Related Work}
\label{sec:related-work}

AutoML is a topic that has been intensively studied over the last decade. 
We briefly summarize related work in this section and readers can consult latest surveys~\cite{automl_book,automl,DBLP:journals/corr/abs-1904-12054,he2020automl} for more details.

\vspace{-0.5em}
\paragraph*{End-to-End AutoML}
End-to-end AutoML, the focus of this work, aims to automate the development process of the end-to-end ML pipeline, including feature preprocessing, feature engineering, algorithm selection, and hyper-parameter tuning.
Often, this is modeled as a black-box optimization problem~\cite{Hutter2015} and solved jointly~\cite{feurer2015efficient,ThoHutHooLey13-AutoWEKA,olson2019tpot}.
Apart from grid search and random search~\cite{bergstra2012random}, genetic programming~\cite{Mohr2018,olson2019tpot} and Bayesian optimization (BO)~\cite{bergstra2011algorithms,hutter2011sequential,snoek2012practical,eggensperger2013towards,bo_survey} has become prevailing frameworks for this problem. One challenge of end-to-end AutoML is the staggeringly huge search space that one has to support and many of these methods suffer from scalability issues.
In addition, meta-learning~\cite{DBLP:journals/corr/abs-1810-03548} systematically investigates the interactions that different ML approaches perform on a wide range of learning tasks, and then learns from this experience, to accomplish new tasks much faster.
Several meta-learning approaches~\cite{DeSa2017,Hutter2014,VanRijn2018,feurer2015efficient} can guide ML practitioners to design better search spaces for AutoML tasks.

Many end-to-end AutoML systems have raised the abstraction level of ML.
\texttt{auto-weka}~\cite{ThoHutHooLey13-AutoWEKA}, \texttt{hyperopt-sklearn}~\cite{komer2014hyperopt}  and \texttt{auto-sklearn}~\cite{feurer2015efficient} are the main representatives of BO-based AutoML systems.
\texttt{auto-sklearn} is one of the most popular open-source framework.
\texttt{TPOT}~\cite{olson2019tpot} and ML-Plan~\cite{Mohr2018} use genetic algorithm and hierarchical task networks planning respectively to optimize over the pipeline space, and require discretization of the hyper-parameter space.
AlphaD3M~\cite{Drori2018} integrates reinforcement learning with Monte Carlo tree search (MCTS) to solve AutoML problems but without imposing efficient decomposition over hyper-parameters and algorithm selection. AutoStacker~\cite{Chen2018} focuses on ensembling and cascading to generate complex pipelines, and solves the CASH problem~\cite{feurer2015efficient} via random search.
Furthermore, a growing number of commercial enterprises also export their AutoML services to their users, e.g., \texttt{H2O}~\cite{ledell2020h2o}, Microsoft's Azure Machine Learning~\cite{barnes2015azure}, Google's Prediction API~\cite{GooPre}, Amazon Machine Learning~\cite{liberty2020elastic} and IBM’s Watson Studio AutoAI~\cite{IBMc}.

\vspace{-0.5em}
\paragraph*{Automating Individual Components}
Apart from end-to-end AutoML, many efforts have been devoted to studying sub-problems in AutoML: (1) feature engineering~\cite{Khurana2016,Kaul2017,Katz2017,Nargesian2017,Khurana2018}, (2) algorithm selection~\cite{ThoHutHooLey13-AutoWEKA,komer2014hyperopt,feurer2015efficient,efimova2017fast,liu2019automated,li2020efficient}, and (3) hyper-parameter tuning~\cite{hutter2011sequential,snoek2012practical,bergstra2011algorithms,li2018hyperband,jamieson2016non,falkner2018bohb,li2020mfeshb,swersky2013multi,kleinfbhh17,kandasamy2017multi,poloczek2017multi,hu2019multi,sen2018noisy,wu2019practical}. 
Meta-learning methods~\cite{wistuba2016two,golovin2017google,feurer2018scalable} for hyper-parameter tuning can leverage auxiliary knowledge acquired from previous tasks to achieve faster optimization.
Several systems offer a subset of functionalities in the end-to-end process. 
Microsoft's NNI~\cite{nni} helps users to automate feature engineering, hyper-parameter tuning, and model compression.
Recent work~\cite{liu2019automated} leverages the ADMM optimization framework to decompose the CASH problem~\cite{feurer2015efficient}, and solves two easier sub-problems.
Berkeley's Ray project~\cite{Moritz2007} provides the \texttt{tune} module~\cite{liaw2018tune} to support scalable hyper-parameter tuning tasks in a distributed environment. 
Featuretools~\cite{kanter2015deep} is a Python library for automatic feature engineering.
Unlike these works, in this paper, we focus on deriving an end-to-end solution to the AutoML problem, where the sub-problems are solved in a joint manner.

\vspace{-0.5em}
\section{\sys and Building Blocks}
\label{sec:building-blocks}

The goal of \sys is to enable scalability with respect to the underlying AutoML search
space. 
As a result, its design focuses on the \emph{decomposition} of a given search space.
In this section, we first introduce key building blocks in \sys, and in Section~\ref{sec:exec-plan} we describe how multiple building blocks are put together to compose a \sys \emph{execution plan} in a modular way.
Later in Section 5, we introduce additional optimizations for these building blocks.

\vspace{-0.5em}
\subsection{Search Space of End-to-End AutoML}
\label{sec:building-blocks:search-space}

We describe the search space of end-to-end AutoML 
following \texttt{auto-sklearn}.
The input to the system is a dataset $D$, containing a set of training samples. 
The user also provides a pre-defined metric, e.g., validation accuracy or cross-validation accuracy, to measure the \textit{utility} of a given ML pipeline.
The output of an end-to-end AutoML system is an ML pipeline that achieves good utility.

To find such an ML pipeline, the system searches over a large search space of possible
pipelines and picks one that maximizes the pre-defined utility. 
This search space is a composition of (1) feature engineering operators, (2) ML algorithms/models, and (3) hyper-parameters. 

\vspace{-0.5em}
\paragraph*{\underline{Feature Engineering}} 
The feature engineering process takes as input a dataset $D$ and outputs a new dataset $D'$.
It achieves this by transforming the input dataset via a set of data transformations.
In \texttt{auto-sklearn}, it further defines multiple \textit{stages} of the feature engineering process: (1) \textit{preprocessing}, (2) \textit{rescaling}, (3) \textit{balancing}, and (4) \textit{feature\_transforming}.
For each stage, the system chooses a single transformation to apply. 
For example, for \textit{feature\_transforming}, the system can choose among \texttt{no\_processing}, \texttt{kernel\_pca}, \texttt{polynomial}, \texttt{select\_percentile}, etc.

\vspace{-0.5em}
\paragraph*{\underline{ML Algorithms}} 
Given a transformed dataset $D'$, the system then picks an ML algorithm to train. Since different ML algorithms are suitable for different types of tasks, the system needs to consider a diverse range of possible ML algorithms.
Taking \texttt{auto-sklearn} as an example, the search space for ML algorithms contains \texttt{Linear\_Model}, \texttt{Support\_Vector\_Machine}, \texttt{Discriminant\_Analysis}, \texttt{Random\_Forest}, etc.

\vspace{-0.5em}
\paragraph*{\underline{Hyper-parameters}}
Each ML algorithm has its own sub-search space for hyper-parameter tuning --- if we choose to use a certain ML algorithm, we also have to specify the corresponding hyper-parameters. 
The hyper-parameters fall into three categories: continuous (e.g., \texttt{sub-sample\_rate} for \texttt{Random\_Forest}), discrete (e.g., \texttt{maximal\_depth} for \texttt{Decision\_Tree}), and categorical (e.g., \texttt{kernel\_type} for \texttt{Lib\_SVM}).  

If the system makes a concrete pick for each of the above decisions, then it can compose a concrete ML pipeline and evaluate its utility. This is often an expensive process
since it involves training an ML model. 
To find the optimal ML pipeline, the system evaluates the utility of different ML pipelines in an iterative manner following a \textit{search strategy}, and picks the one that maximizes the utility.

For example, \texttt{auto-sklearn} handles the above search space \textit{jointly} and optimizes it with Bayesian optimization (BO)~\cite{bo_survey}.
Given an initial set of function evaluations, BO proceeds by fitting a surrogate model to those observations, specifically a \textit{probabilistic Random Forest} in \texttt{auto-sklearn}, and then chooses which ML pipeline to evaluate from the search space by optimizing an acquisition function that balances exploration and exploitation. 

\vspace{-0.5em}
\subsection{Building Blocks}

Unlike \texttt{auto-sklearn}, \sys decomposes the above search space into smaller subspaces.
One interesting design decision in \sys is to introduce a \emph{structured abstraction} to express different \emph{decomposition strategies}.
A decomposition strategy is akin to an \emph{execution plan} in relational database management systems, which is composed of \emph{building blocks} akin to relational operators.
A building block itself can be viewed as an \emph{atomic} decomposition strategy.
We next present the details of the building blocks implemented by \sys, and we will introduce how to use these blocks to compose \sys execution plans in Section~\ref{sec:exec-plan}.

\paragraph*{\underline{Goal}}
The \textit{goal} of \sys is to solve:
\[
\min_{x_1,...,x_n} f(x_1,...,x_n; D),
\]
where $x_1,...,x_n$ is a set of $n$ variables and each of them has domain $\mathbb{D}_{x_i}$
for $i \in [n]$.
Together, these $n$ variables define a search space $(x_1,...,x_n) \in \prod_i \mathbb{D}_{x_i}$.
$D$ corresponds to the input dataset, which is a \textit{set} of input samples. 
In our setting, $f(\cdot)$ is a black-box function that we can only evaluate (but not exploiting the derivative).
Given constant $\bm{c}$ in the composite domain $\bm{c} \in \prod_i \mathbb{D}_{x_i}$, we use the notation
\[
f(\{(x_1,...x_n)=\bm{c}\}; D)
\]
as the value of evaluating $f$ by substituting $(x_1,...x_n)$ with $\bm{c}$.

\paragraph*{\underline{Subgoal}} 
One key decision of \sys is to solve the optimization problem on a search space by decomposing it into multiple smaller subspaces, each of which will be solved by one \textit{building block}. 
We define optimizing over each of these smaller subspaces as a \textit{subgoal} of the original problem.
Formally, a subgoal $g$ is defined by two components: $\bm{\bar{x}}_g \subseteq \{x_1,...x_n\}$ as a subset of variables, and 
$\bm{\bar{c}}_g \in \prod_{x_i \in \bm{\bar{x}}_g} \mathbb{D}_{x_i}$ as an assignment in the domain of all variables in $\bm{\bar{x}}_g$. 
Let $\bm{\bar{x}}_{-g} = \{x_1,...,x_n\} - \bm{\bar{x}}_g$ be all variables that are \textit{not} in $\bm{\bar{x}}_g$.

Each subgoal defines a function $f_g$ over a smaller search space, which is constructed by \textit{substituting} all variables in $\bm{\bar{x}}_g$ with $\bm{\bar{c}}_g$:
\[
f_g = f[\bm{\bar{x}}_g \slash \bm{\bar{c}}_g] : 
{\bm{z}} \in \prod_{x_i \in \bm{\bar{x}}_{-g}} \mathbb{D}_{x_i} \mapsto f(\{\bm{\bar{x}}_g = \bm{\bar{c}}_g; 
\bm{\bar{x}}_{-g}={\bm{z}}\}; D).
\]

\paragraph*{\underline{Building Block}}
Each subgoal $g$ corresponds to one building block $B_{g, D}$, whose goal is
to solve
\[
\min_{\bm{\bar{x}}_{-g}} f_g(\bm{\bar{x}}_{-g}; D).
\]
A building block $B_{g, D}$ imposes several assumptions on $g$ and $D$.
First, given an assignment $\bm{\bar{c}}_{-g}$ to $\bm{\bar{x}}_{-g}$, it is able to evaluate the value of the function $f_g(\bm{\bar{c}}_{-g},D)$.
Note that such an evaluation can often be expensive and \sys tries to minimize the number of times that such a function is evaluated.
Second, given a dataset $D$, a building block has the knowledge about how to subsample a smaller dataset $\tilde{D} \subseteq D$ and then conduct evaluations on such a subset $\bm{x} \mapsto f_g(\bm{x}; \tilde{D})$. 
Third, we assume that the building block has access to a cost model about the cost of an evaluation at $\bm{x}$, $C_{g, D, \bm{x}}$.

\paragraph*{\underline{Interfaces}} 
All implementations of a building block follow an interactive optimization process. A building block exposes several interfaces. 
First, one can initialize a building block via
\[
B_{g, D} \leftarrow \texttt{init}(f, \bm{\bar{x}}_g, \bm{\bar{c}}_g, D),
\]
which creates a building block.
Second, one can query the current best solution found in $B_{g, D}$ by
\[
\bm{\hat{x}} \leftarrow \texttt{get\_current\_best}(B_{g, D}).
\]
Furthermore, one can ask $B_{g, D}$ to iterate once via
\[
\texttt{do\_next!}(B_{g, D}),
\]
where `\texttt{!}' indicates potential change on the state of the input $B_{g, D}$.

Last but not least, one can query a building block about its \emph{expected utility} (EU) if given $K$ more budget units (e.g., seconds) via
\[
[l, u] \leftarrow \texttt{get\_eu}(B_{g, D}, K).
\]
By adopting a similar design principle used in the existing AutoML systems~\cite{feurer2015efficient,olson2019tpot,liu2019automated}, in \sys we estimate EU by \emph{extrapolation} into the ``future'' with more available budget.
Given the inherent uncertainty in our estimation method, rather than returning a single point estimate, we instead return a lower bound $l$ and an upper bound $u$.
We refer readers to~\cite{li2020efficient} for the details of how the lower and upper bounds are established.
Moreover, one can query a building block about its \textit{expected utility improvement} (EUI) via
\[
\delta \leftarrow \texttt{get\_eui}(B_{g, D}).
\]
Note that, different from EU, EUI is the expected \textit{improvement} over the current observed utility if given $K$ more budget units.
In \sys, we estimate EUI by taking the mean of the observed improvements from history, following Levine et al~\cite{levine2017rotting}.

\vspace{-0.5em}
\subsection{\bf Three Types of Building Blocks}

Decomposition is the cornerstone of \sys's design.
Given a search space, apart from exploring it jointly, there are two classical ways of decomposition --- to partition the search space via conditioning on different values of a certain variable (in a similar spirit of \emph{variable elimination}~\cite{dechter1998bucket}), or to decompose the problem into multiple smaller ones by introducing equality constraints (in a similar spirit of \emph{dual decomposition}~\cite{caroe1999dual}). This inspires \sys's design, which supports three types of building blocks: 
(1) \emph{joint block} that simply optimizes the input subspace using Bayesian optimization;
(2) \emph{conditioning block} that further divides the input subspace into smaller ones by conditioning on one particular input variable;
and (3) \emph{alternating block} that partitions the input subspace into two and optimizes each one \emph{alternately}.
Note that both \emph{conditioning block} and \emph{alternating block} would generate new building blocks with smaller subgoals.
We next present the implementation details for each type of building block.

\vspace{-0.5em}
\subsubsection{Joint Block}



A joint block directly optimizes its subgoal via Bayesian optimization (BO)~\cite{bo_survey}.
Specifically, BO based method - SMAC~\cite{hutter2011sequential} has been used by many applications where evaluating the objective function is computationally expensive.
It constructs a probabilistic surrogate model $M$ to capture the relationship between the input variables $\bm{\bar{x}}$ and the objective function value $\psi$, and refines $M$ iteratively using past observations $(\bm{\bar{x}}, \psi)$s.

The implementation of \texttt{do\_next!} for a joint block consists of the following three steps:
\vspace{-0.2em}
\begin{enumerate}
    \item Use the surrogate model $M$ to select $\bm{\bar{x}}$ that maximizes an acquisition function. In our implementation, we use \emph{expected improvement} (EI)~\cite{jones1998efficient} as the acquisition function, which has been widely used in BO community.
    \item Evaluate the selected $\bm{\bar{x}}$ and obtain its value about the objective function (i.e., the subgoal) $\psi=f_g(\bm{\bar{x}})+\epsilon$ with $\epsilon \sim \mathcal{N}(0, \sigma^2)$, where $\mathcal{N}$ is the normal distribution.
    \item Refit the surrogate model $M$ on the observed $(\bm{\bar{x}}, \psi)$s.
\end{enumerate}

\textbf{Early-Stopping based Optimization.}
For large datasets, early-stopping based methods, e.g., Successive Halving~\cite{jamieson2016non}, Hyperband~\cite{li2018hyperband}, BOHB~\cite{falkner2018bohb}, MFES-HB~\cite{li2020mfeshb}, etc, can terminate the evaluations of poorly-performed configurations in advance, thus speeding up the evaluations. 
\sys supports MFES-HB~\cite{li2020mfeshb}, which combines the benefits of Hyperband and Multi-fidelity BO~\cite{Wu2019,takeno2020multifidelity}, to optimize a joint block, in addition to vanilla BO.

\begin{algorithm}[t]
  \scriptsize
  \SetAlgoLined
  \KwIn{A conditioning block $B_{g, D}$, budget $K$.}
  \SetAlgoLined
  \caption{The \texttt{do\_next!} of conditioning block}
  \label{alg:cond:next}
  Let $B_1$, ..., $B_m$ be all active (have not been eliminated) child blocks\;
  \For{$1\leq i\leq L$}
  {
    \For{$1\leq j\leq m$}
    {
        \texttt{do\_next!}$(B_j)$\;
    }
  }
  \For{$1\leq j\leq m$}
  {
    $[l_j, u_j]\leftarrow$~\texttt{get\_eu}$(B_j, K)$\;
  }
  Eliminate child blocks that are \emph{dominated} by others, using $[l_j, u_j]$ for $1\leq j\leq m$\;
\end{algorithm}

\vspace{-0.5em}
\subsubsection{Conditioning Block}
A conditioning block decomposes its input $\bm{\bar{x}}$ into $\bm{\bar{x}} = \{x_c\} \cup \bm{\bar{y}}$, where $x_c$ is a single variable with domain $\mathbb{D}_{x_c}$.
It then creates one new building block for each possible value $d \in \mathbb{D}_{x_c}$ of $x_c$:
\[
\min_{\bm{\bar{y}}} g_d(\bm{\bar{y}}; D) \equiv f(\{x_c=d, \bm{\bar{y}}\}; D).
\]
As a result, $|\mathbb{D}_{x_c}|$ new (child) building blocks are created.

The conditioning block aims to identify optimal value for $x_c$, and many previous AutoML researches have used Bandit algorithms for this purpose~\cite{liu2019automated,jamieson2016non,yangli2020,li2020mfeshb}. 
In \sys, we follow these previous work and 
model it as a multi-armed bandit (MAB) problem, while our framework is flexible enough to incorporate other algorithms when they are available.
There are $|\mathbb{D}_{x_c}|$ arms, where each arm corresponds to a child block.
Playing an arm means invoking the \texttt{do\_next!} primitive of the corresponding child block.

Algorithm~\ref{alg:cond:next} illustrates the implementation of \texttt{do\_next!} for a conditioning block.
It starts by playing each arm $L$ times in a Round-Robin fashion (lines 2 to 4).
Here, $L$ is a user-specified configuration parameter of \sys.
In our current implementation, we set $L=5$.
We then obtain the lower and upper bounds of the expected utility of each child block by invoking its \texttt{get\_eu} primitive (lines 5 to 6), and eliminate child blocks that are dominated by others (line 7).
The elimination works as follows.
Consider two blocks $B_i$ and $B_j$: if the upper bound $u_i$ of $B_i$ is less than the lower bound $l_j$ of $B_j$, then the block $B_i$ is eliminated.
An eliminated arm/block will not be played in future invocations of \texttt{do\_next!}.

\begin{algorithm}[t!]
  \scriptsize
  \SetAlgoLined
  \KwIn{An alternating block $B_{g, D}$ with search space $\bm{\bar{x}}=\bm{\bar{y}} \cup \bm{\bar{z}}$.}
  \SetAlgoLined
  \caption{The \texttt{init} of alternating block}
  \label{alg:alternating:init}
  Initialize $\bm{\bar{y}}$ and $\bm{\bar{z}}$ with default values $\bm{\bar{y}}_0$ and $\bm{\bar{z}}_0$\;
  $B_1\leftarrow$~\texttt{init}$(f, \bm{\bar{z}}, \bm{\bar{z}}_0, D)$\;
  $B_2\leftarrow$~\texttt{init}$(f, \bm{\bar{y}}, \bm{\bar{y}}_0, D)$\;
  \For{$1\leq i\leq L$}
  {
    \texttt{do\_next}$(B_1)$\;
    $\bm{\bar{y}}_i\leftarrow$~\texttt{get\_current\_best}$(B_1)$\;
    \texttt{set\_var}$(B_2, \bm{\bar{y}}, \bm{\bar{y}}_i)$\;
    \texttt{do\_next}$(B_2)$\;
    $\bm{\bar{z}}_i\leftarrow$~\texttt{get\_current\_best}$(B_2)$\;
    \texttt{set\_var}$(B_1, \bm{\bar{z}}, \bm{\bar{z}}_i$)\;
  }
\end{algorithm}

\noindent
\textbf{Remark:} We have simplified the above elimination criterion by using the lower and upper bounds calculated given $K$ budget units for \emph{each arm}.
In fact, these $K$ budget units are \emph{shared} by all the arms, and as a result, each arm actually has fewer budget units than $K$.
Our assumption is that, $K$ is sufficiently large so that one can play \emph{all arms} until (the observed distribution of rewards of) \emph{each arm} converges.
Otherwise, the lower and upper bounds obtained may be \emph{over-optimistic}, and as a result, may lead to incorrect eliminations.
Fortunately, our assumption usually holds in practice, where arms converge relatively fast.

\vspace{-0.5em}
\subsubsection{Alternating Block}

An alternating block decomposes its input search space into $\bm{\bar{x}} = \bm{\bar{y}} \cup \bm{\bar{z}}$, and explores $\bm{\bar{y}}$ and $\bm{\bar{z}}$ in an \emph{alternating} way.
Similarly, we also model the optimization in alternating block as an MAB problem.
Algorithm~\ref{alg:alternating:init} illustrates how its \texttt{init} primitive works.
It first creates two child blocks $B_1$ and $B_2$, which will focus on optimizing for $\bm{\bar{y}}$ and $\bm{\bar{z}}$ respectively (lines 1 to 3).
It then (again) views $B_1$ and $B_2$ as two arms and plays them using Round-Robin (lines 4 to 10).
Note that, when $B_1$ optimizes $\bm{\bar{y}}$ (resp. when $B_2$ optimizes $\bm{\bar{z}}$), it uses the current best $\bm{\bar{z}}$ found by $B_2$ (resp. the current best $\bm{\bar{y}}$ found by $B_1$).
This is done by the \texttt{set\_var} primitive (invoked at line 7 for $B_2$ and line 10 for $B_1$).

One problem of our alternating MAB formulation is that the utility improvements of the two building blocks often vary dramatically in practice.
For example, some applications are very sensitive to the features being used (e.g., normalized vs. non-normalized features) while hyper-parameter tuning will offer little or even no improvement.
In this case, we should spend more resources on looking for good features instead of tuning hyper-parameters.
Our key observation is that, the \emph{expected utility improvement} (EUI) decays as optimization proceeds.
As a result, we propose to use EUI as an indicator that measures the \emph{potential} of pulling an arm further.
Algorithm~\ref{alg:alternating:next} illustrates the details of this idea when used to implement the \texttt{do\_next!} primitive.

\begin{algorithm}[t]
  \scriptsize
  \SetAlgoLined
  \KwIn{An alternating block $B_{g, D}$ with budget $K$.}
  \SetAlgoLined
  \caption{The \texttt{do\_next!} of alternating block}
  \label{alg:alternating:next}
  $\delta_1\leftarrow$~\texttt{get\_eui}$(B_1)$\;
  $\delta_2\leftarrow$~\texttt{get\_eui}$(B_2)$\;
  \If{$\delta_1\geq \delta_2$}
  {
    $\bm{\bar{z}}_{\text{best}}\leftarrow$~\texttt{get\_current\_best}$(B_2)$\;
    \texttt{set\_var}$(B_1, \bm{\bar{z}}, \bm{\bar{z}}_{\text{best}})$\;
    \texttt{do\_next}$(B_1)$\;
  }\Else{
    $\bm{\bar{y}}_{\text{best}}\leftarrow$~\texttt{get\_current\_best}$(B_1)$\;
    \texttt{set\_var}$(B_2, \bm{\bar{y}}, \bm{\bar{y}}_{\text{best}})$\;
    \texttt{do\_next}$(B_2)$\;
  }
\end{algorithm}

Specifically, Algorithm~\ref{alg:alternating:next} starts by polling the EUI of both child blocks (lines 1 and 2).
Recall that the EUI is estimated by taking the mean of historic observations.
It then compares the EUIs and picks the arm/block with larger EUI to play next (lines 3 to 10).
Before pulling the winner arm, again it will use the current best settings found by the other arm/block (lines 4 to 6, lines 8 to 10).

\vspace{-0.5em}
\subsubsection{Discussion: Pros and Cons of Building Blocks} 
While the joint block is the most straightforward way to solve the optimization problem associated, it is difficult to scale Bayesian optimization to a large search space~\cite{wang2013bayesian,li2020efficient}. 
The alternating block addresses this scalability issue by decomposing the search space into two smaller subspaces, though with the assumption that the improvements of the two subspaces are \emph{conditionally independent} of each other.
As a result, the alternating block is a better choice when such an assumption approximately holds.
The conditioning block is capable of pruning the search space \emph{as optimization proceeds}, when bad arms are pulled less often or will not be played anymore, with the limitation that it can only work for conditional variables that are \emph{categorical}. 
For non-categorical variables, one possible way to use conditioning blocks is to split the value range of variables. 
For example, given a numerical variable that ranges from 1 to 3, we split it into two ranges, which are [1, 2) and [2, 3). 
During the optimization iteration, we first choose one sub-range and then optimize the splitted space along with its corresponding subspace.

In addition, \sys uses bandit-based algorithms from the existing literature~\cite{levine2017rotting,li2020efficient} as default in both the alternating and conditioning block, and other bandit-based algorithms, such as successive halving~\cite{jamieson2016non}, Hyperband~\cite{li2018hyperband},  BOHB~\cite{falkner2018bohb} and MFES-HB~\cite{li2020mfeshb}, can also be used in these blocks.

\vspace{-0.5em}
\subsubsection{Discussion: Comparing Different Building Blocks}
Joint blocks are the default blocks that can be applied to all problems. 
When the search space is rather large, conditioning and alternating blocks can be helpful.
If the search space contains a categorical hyper-parameter, under which the subspace of each choice is conditionally independent with each other, 
the conditioning block can be used instead of exploring the entire space. 
If the search space can be decomposed into two approximately independent subspaces, the alternating block can be applied to this case.
As a result, a scalable system needs to be able to decompose the problem in different ways and pick the most suitable building blocks. 
This forms a \sys execution plan, which we will describe in the next section.
In Section~\ref{sec:exec-plan}, we explore the possibility of automatically choosing building blocks to use by maximizing the empirical accuracy of different execution plans, given a pre-defined set of datasets.

\vspace{-0.5em}
\section{\sys Execution Plan}
\label{sec:exec-plan}

Given a pre-defined search space, the input of \sys is (1) a dataset $D$, (2) a utility metric (e.g, cross-validation accuracy) which defines the objective function $f$, and (3) a time budget.
\sys then decomposes a large search space into an \emph{execution plan}, following some specific \emph{decomposition strategy}.

\begin{figure}
\centering
\vspace{-1em}
\includegraphics[width=0.3\textwidth]{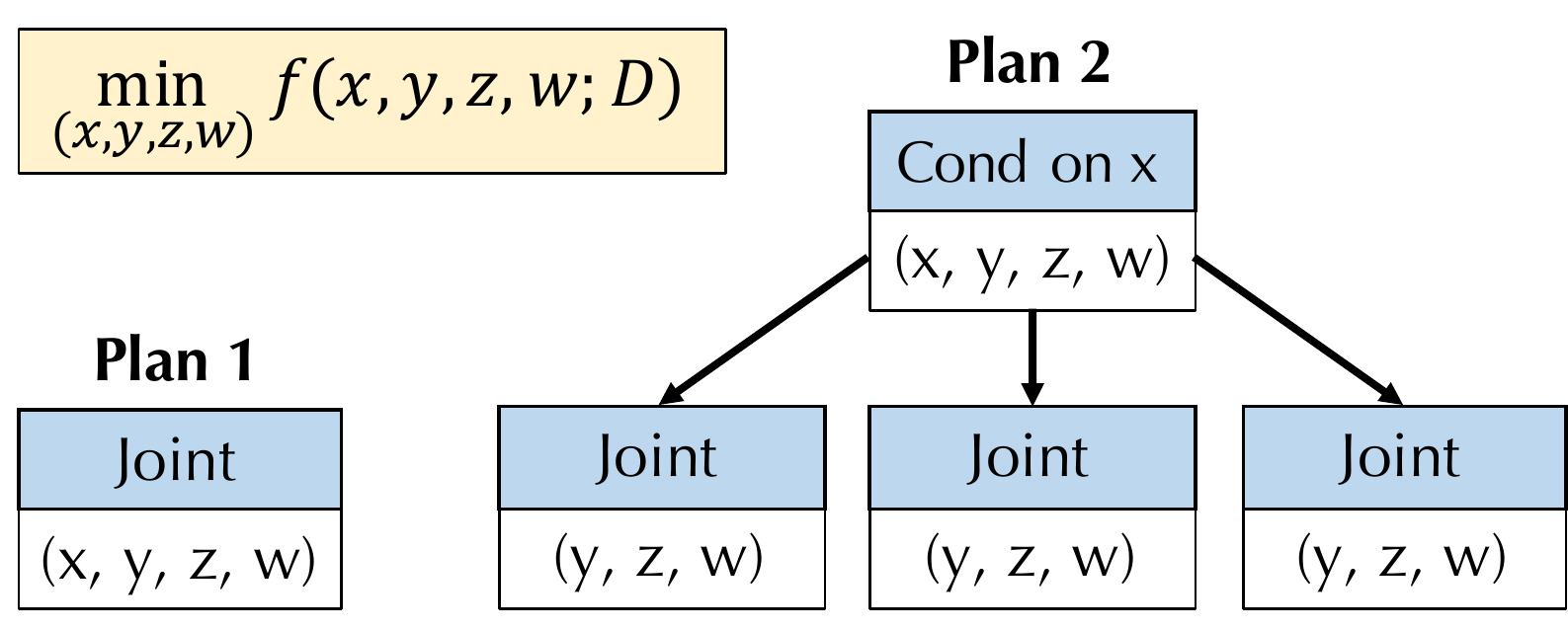}
\vspace{-1em}
\caption{Two different execution plans
for the same optimization problem. Each plan
corresponds to a different way to decompose 
the same search space $(x,y,z,w)$.}
\label{fig:plan}
\vspace{-2em}
\end{figure}

\vspace{-0.5em}
\paragraph*{\underline{\sys Execution Plan}}
Due to space limitation, we omit the formal definition of a \sys execution plan. 
Intuitively, a \sys execution plan is a \textit{tree} of building blocks.
The root node corresponds to a building block solving the problem $f$ with the entire search space, which can be further decomposed into multiple building blocks if necessary, as previously described.
As an example, Figure~\ref{fig:plan} illustrates two possible execution plans
for $f(x,y,z,w;D)$.
\textbf{Plan 1} contains only a single root building block as a joint block, whereas \textbf{Plan 2} first introduces a conditioning block on $x$, and then creates one lower level of building blocks for each possible value of $x$ (in Figure~\ref{fig:plan}, we assume that $|\mathbb{D}_x|=3$).

\vspace{-0.5em}
\paragraph*{\underline{\sys Execution Model}}
To execute a \sys execution plan, we follow a Volcano-style execution that is similar to a relational database~\cite{Graefe1994} --- the system invokes the $\texttt{do\_next!}$ of the root node, which then invokes the $\texttt{do\_next!}$ of one of its child nodes, propagating until the leaf node.
At any time, one can invoke the \texttt{get\_current\_best} of the root node, which returns the current best solution for the entire search space.

\begin{figure}
\centering
\includegraphics[width=0.32\textwidth]{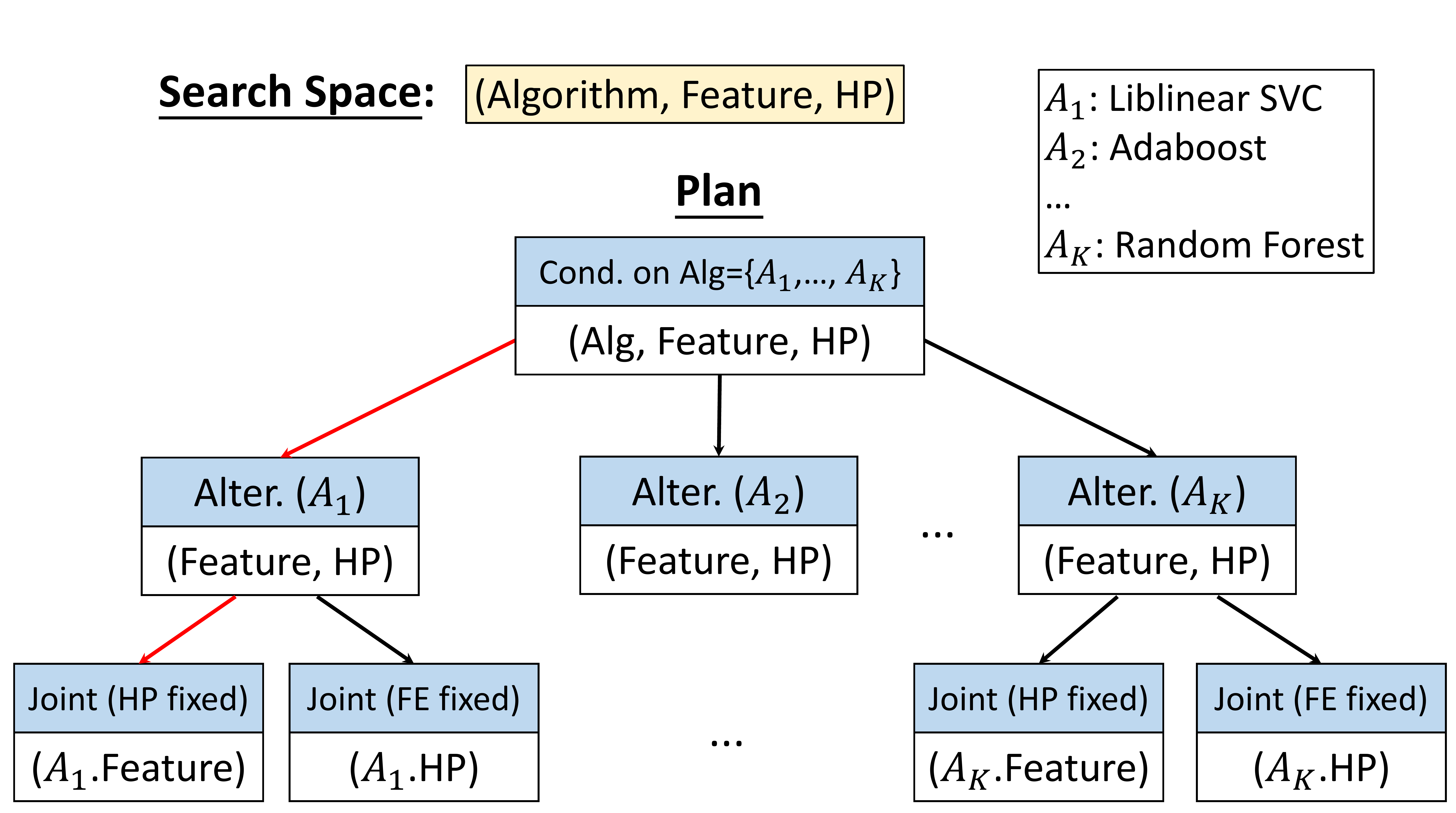}
\vspace{-1em}
\caption{\sys's execution plan for the same search space as explored by \texttt{auto-sklearn}. Here `Alg' and `HP' correspond to Algorithm and hyper-parameters respectively.}
\label{fig:plan_revised}
\end{figure}

\vspace{-0.5em}
\paragraph*{\underline{\sys Plan for \texttt{auto-sklearn}}}
Figure~\ref{fig:plan_revised} presents a \sys execution plan for the same search space explored by \texttt{auto-sklearn}, which consists of the joint search of \textit{algorithms}, \textit{features}, and \textit{hyper-parameters}. 
Instead of conducting the search process in a single joint block, as was done by \texttt{auto-sklearn}, \sys first decomposes the search space via a conditioning block 
on \textit{algorithms} --- this introduces a MAB problem in which each arm corresponds to one particular algorithm.
It then further decomposes each of the conditioned subspaces via an alternating block between feature engineering and hyper-parameter tuning. 
The whole subspace of feature engineering (resp. that of hyper-parameter tuning) is optimized by a joint block.

Concretely, Figure \ref{fig:plan_revised} shows a search space for AutoML with $K$ choices of ML algorithms.
During each iteration, starting from the root node, \sys selects the child node to optimize until it reaches a leaf node, and then optimizes over the subspace in the leaf node. 
As shown by the red lines in Figure \ref{fig:plan_revised}, in this iteration, \sys only tunes the feature engineering pipeline of algorithm $A_1$ while fixing its algorithm hyper-parameters.

\vspace{-0.5em}
\paragraph*{\underline{Alternative Execution Plans}}
Note that the execution plan in Figure~\ref{fig:plan_revised} is not the only possible one.
Our flexible and scalable framework in \sys allows us to explore different execution plans before reaching the proposed one. We enumerate five possible plans in a coarse-grained level, and the results show that the proposed plan performs best. 
The reason why we choose this plan is due to the fundamental property of the AutoML search space --- we observe that, the optimal choices of \textit{features} are different across \textit{algorithms}, which implies that we can first decompose the search space along \textit{ML algorithms}.
The improvements introduced by feature engineering and hyper-parameter tuning are largely complementary, and thus we can optimize them \emph{alternately}.
For feature engineering (resp. hyper-parameter tuning), the subspace is small enough to be handled by a single joint block efficiently.

\begin{figure}
\centering
\includegraphics[width=0.3\textwidth]{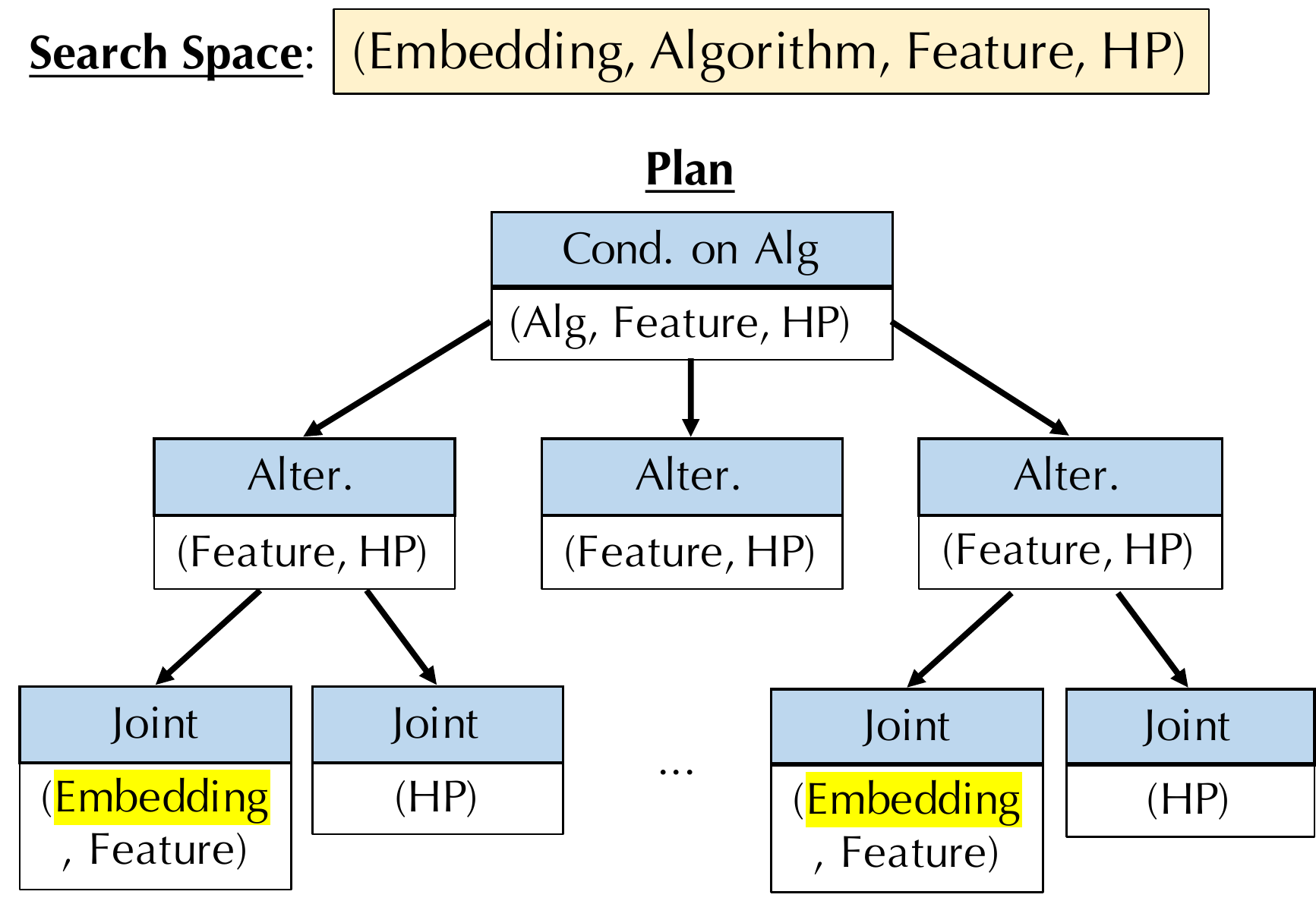}
\vspace{-1em}
\caption{\sys's execution plan for a larger search space enriched by an additional 
embedding selection stage.}
\label{fig:plan_embedding}
\end{figure}

\vspace{-0.5em}
\paragraph*{\underline{\sys Plan for Enriched Search Space}}
We can easily extend \sys and enable functionalities that are not supported by most AutoML systems.
For example, Figure~\ref{fig:plan_embedding} illustrates an execution plan for a search space with an additional stage --- \emph{embedding selection}. Given an input, e.g., image or text, we first choose embeddings based on a collection of TensorFlow Hub pre-trained models, and then conduct algorithm selection, feature engineering, and hyper-parameter tuning. 
We use an execution plan as illustrated in Figure~\ref{fig:plan_embedding}, having the embedding selection step jointly optimized together with the feature engineering. 

\vspace{0.5em}
\noindent
{\bf Discussion: Automatic Plan Generation.}
In principle, the design of \sys opens up the opportunity for ``automatic plan generation'' --- given a collection of benchmark datasets, one could automatically search for the best decomposition strategy of the search space and come up with a physical plan automatically.
While the full treatment of this problem is 
beyond the scope of this paper, we illustrate
the possibility with a very simple strategy.
We automatically enumerate all possible execution plans in a coarse-grained level, and find that our manually specified execution plan in Figure~\ref{fig:plan_revised} outperforms the alternatives. 
There is still an open question that whether we can support finer-grained partition of the search space (e.g., different plans for different subspace of features), and moreover, whether we can conduct efficient automatic plan optimization without enumerating all possible plans. 
These are exciting future directions and we expect 
the endeavor to be non-trivial. 
We hope that this paper sets the ground for this line of research
in the future (e.g., rule-based heuristics or reinforcement learning).


\noindent
{\bf Further Optimization with Meta-learning.}
\label{sec:additional_optimization}
\sys supports meta-learning based techniques --- given
previous runs of the system over similar workloads, to transfer the knowledge and better help the workload at hand --- to accelerate the optimization process of building blocks. Appendix contains the details.


\vspace{-0.5em}
\section{Experimental Evaluation}
\label{sec:experiments}

We compare \sys with state-of-the-art AutoML systems. In our evaluation, we focus on three perspectives: (1) the \emph{performance} of \sys given the \emph{same} search space explored by existing systems, (2) the \emph{scalability} of \sys given larger search spaces, and (3) the \emph{extensibity} of \sys to integrate new components into the search space of AutoML pipelines.


\begin{figure*}
	\centering
	\subfigure[\sys vs. AUSK on CLS]{
		\scalebox{0.27}[0.27]{
			\includegraphics[width=1\linewidth]{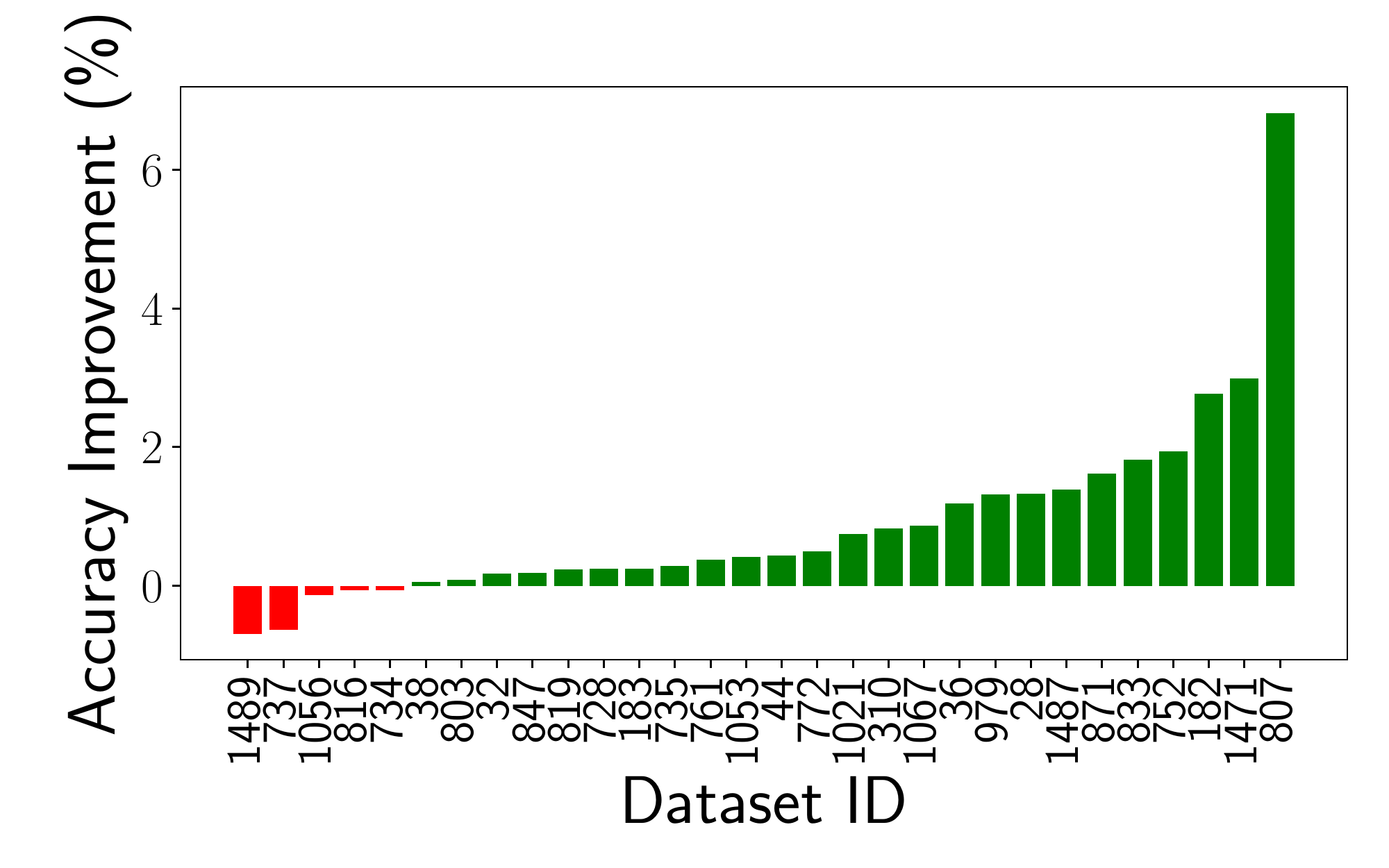}
	}}
	\subfigure[\sys vs. AUSK on REG]{
		\scalebox{0.21}[0.21]{
			\includegraphics[width=1\linewidth]{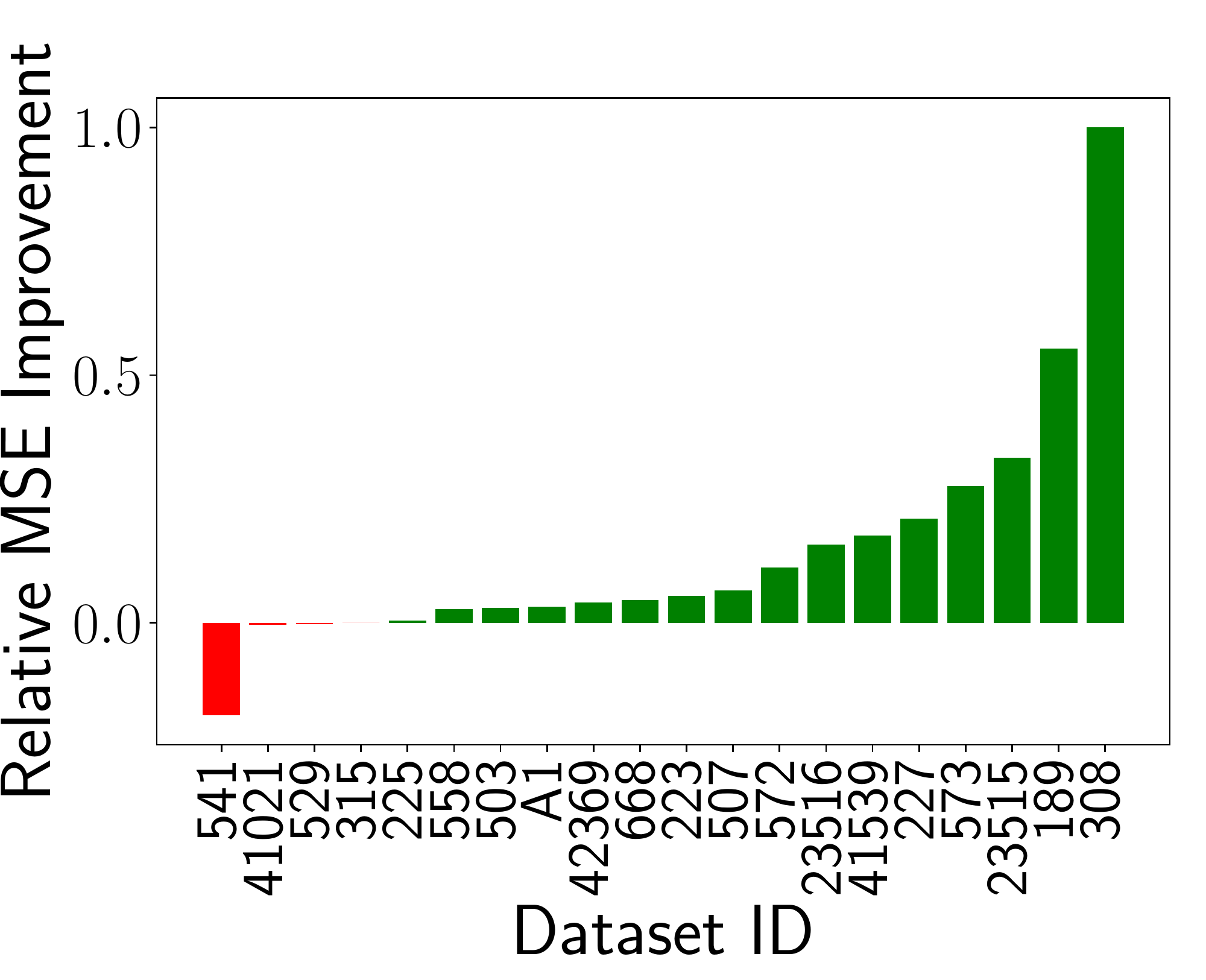}
	}}
	\subfigure[\sys vs. TPOT on CLS]{
		\scalebox{0.27}[0.27]{
			\includegraphics[width=1\linewidth]{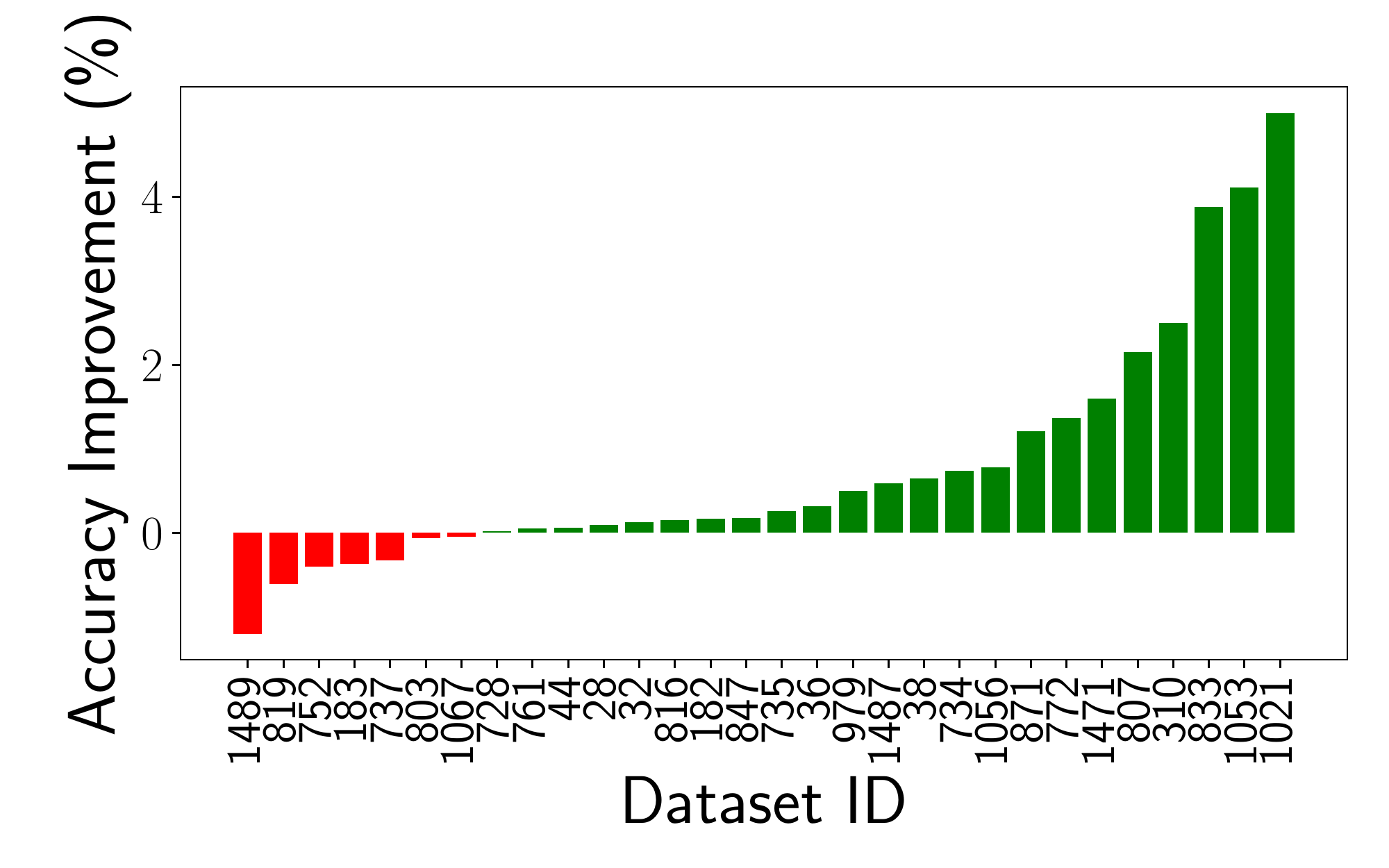}
	}}
    \subfigure[\sys vs. TPOT on REG]{
		\scalebox{0.21}[0.21]{
			\includegraphics[width=1\linewidth]{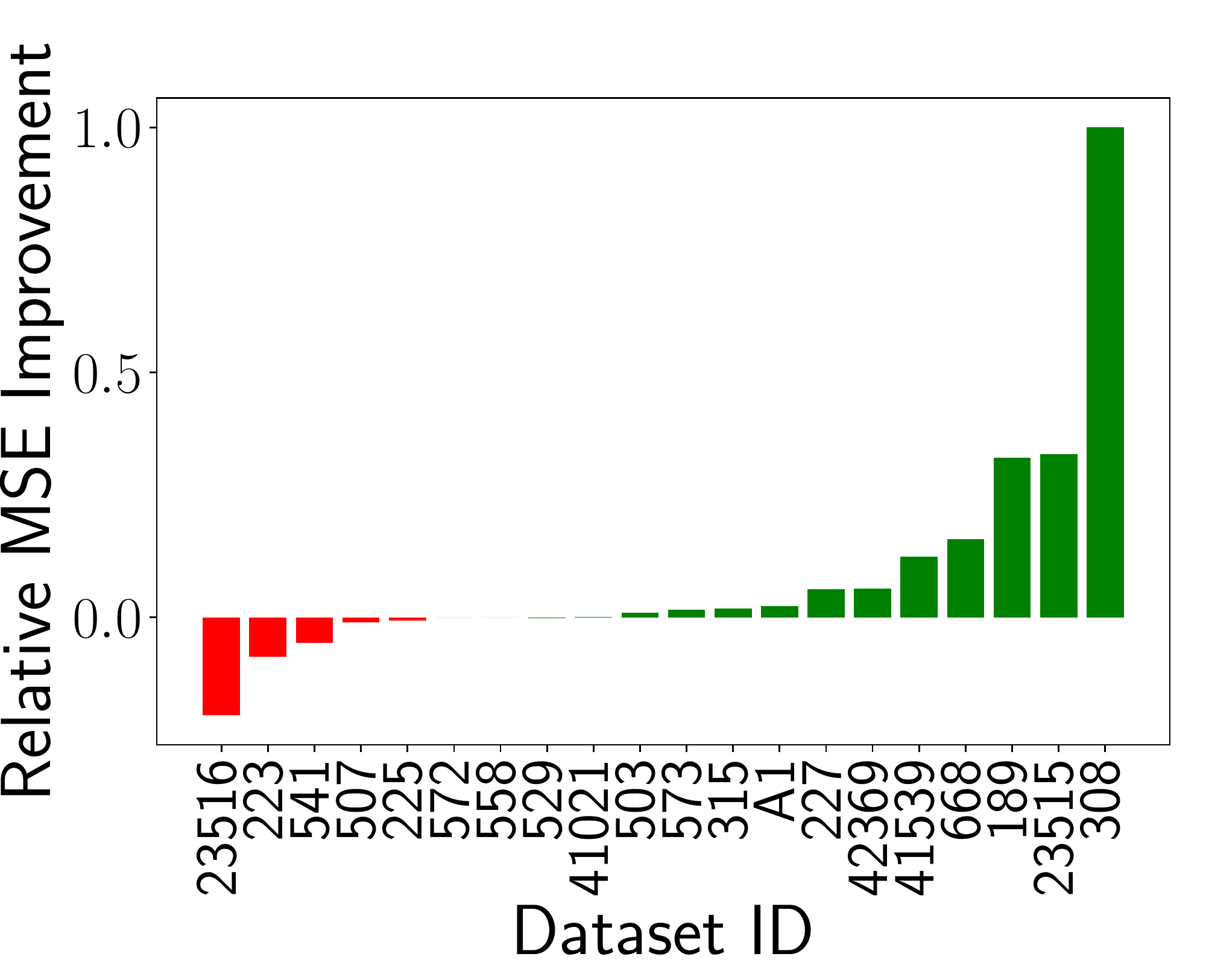}
	}}
	\vspace{-2em}
	\caption{End-to-End results on 30 OpenML classification (CLS) datasets and 20 OpenML regression (REG) datasets.}
	\vspace{-1.5em}
  \label{end2end_cmp}
\end{figure*}

\begin{figure*}
	\centering
	\subfigure[Mnist\_784]{
		\scalebox{0.24}{
			\includegraphics[width=1\linewidth]{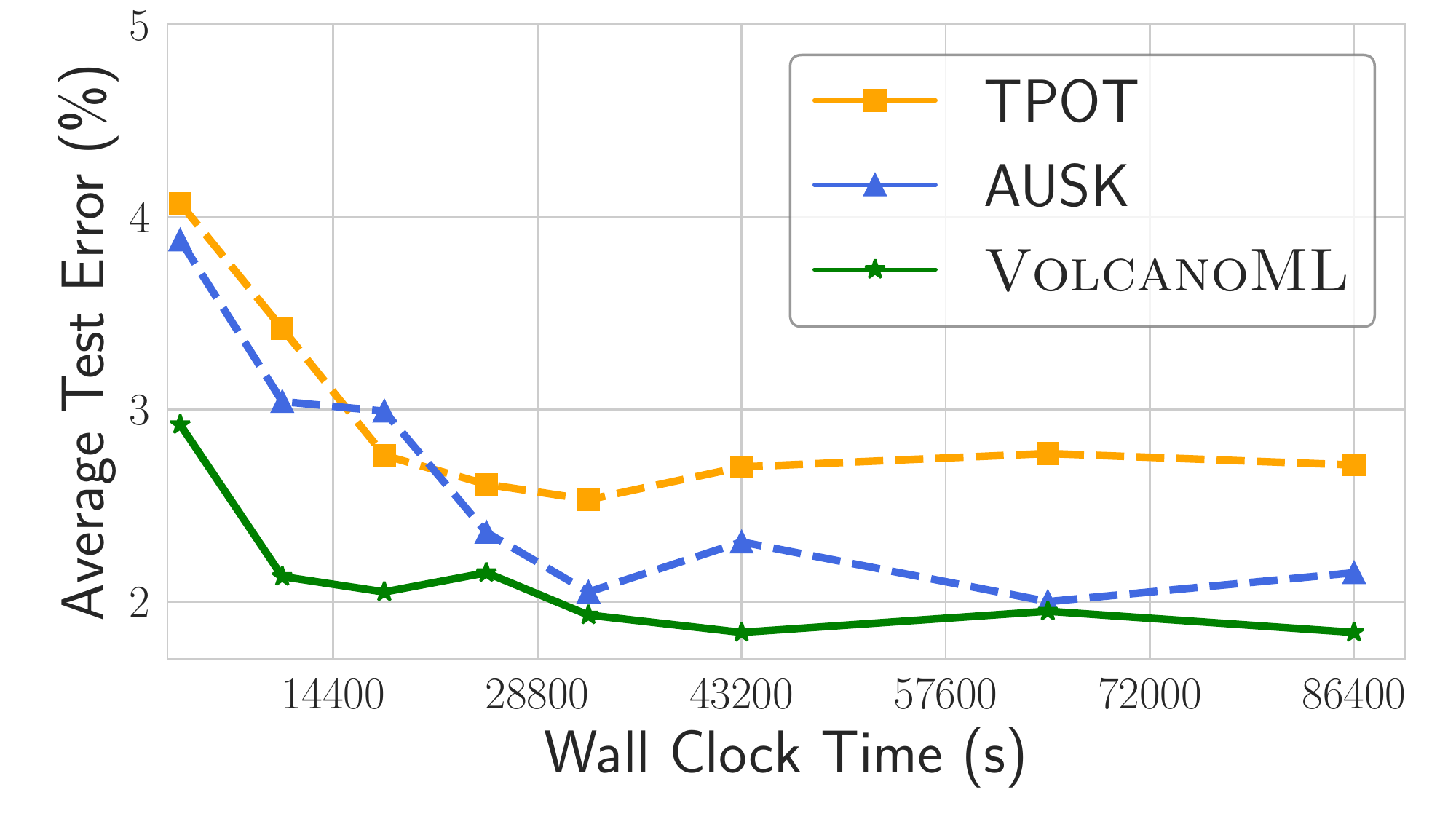}
	}}
	\subfigure[Kropt]{
		\scalebox{0.24}{
			\includegraphics[width=1\linewidth]{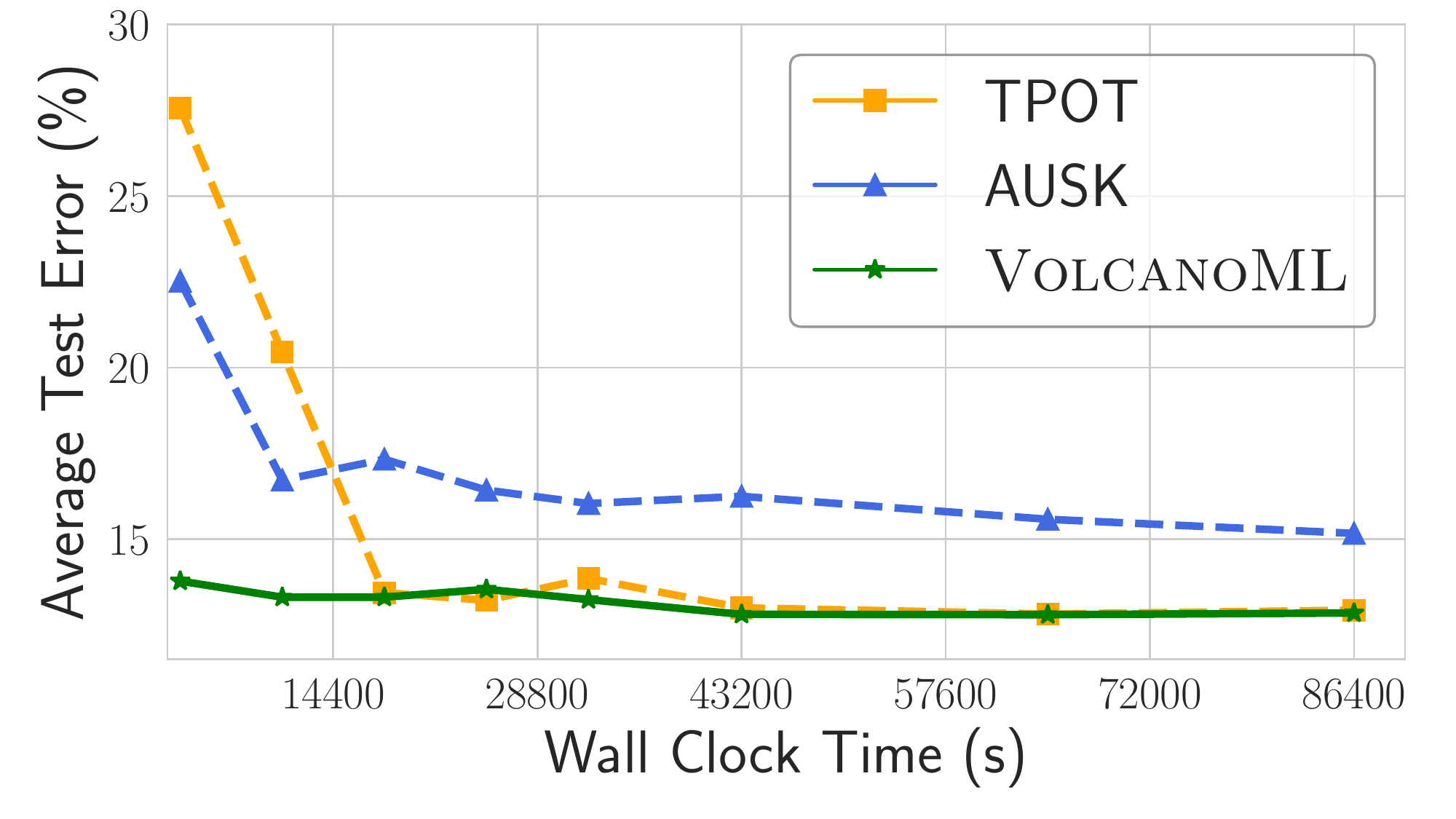}
	}}
	\subfigure[Electricity]{
		\scalebox{0.24}{
			\includegraphics[width=1\linewidth]{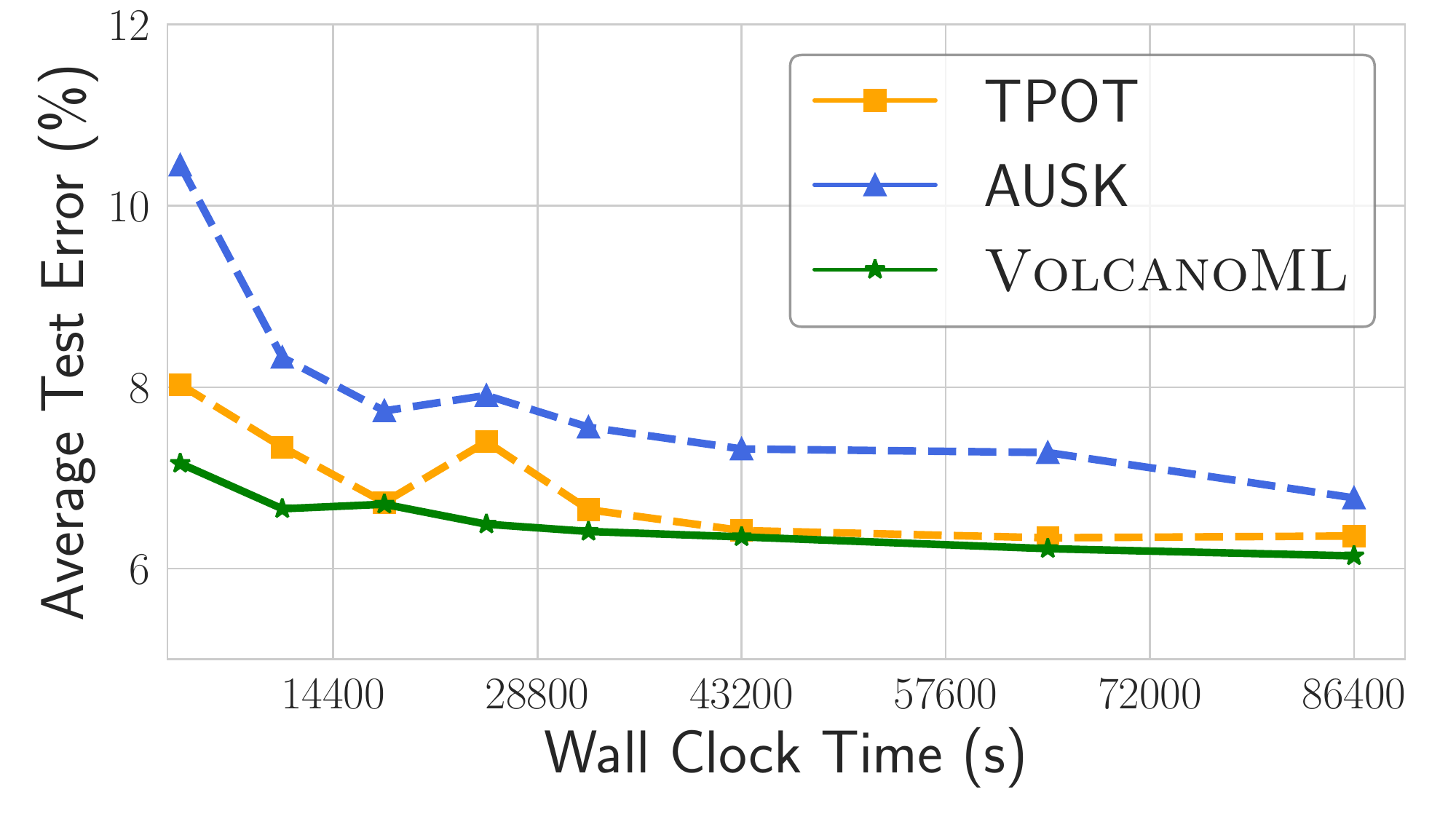}
	}}
    \subfigure[Higgs]{
		\scalebox{0.24}{
			\includegraphics[width=1\linewidth]{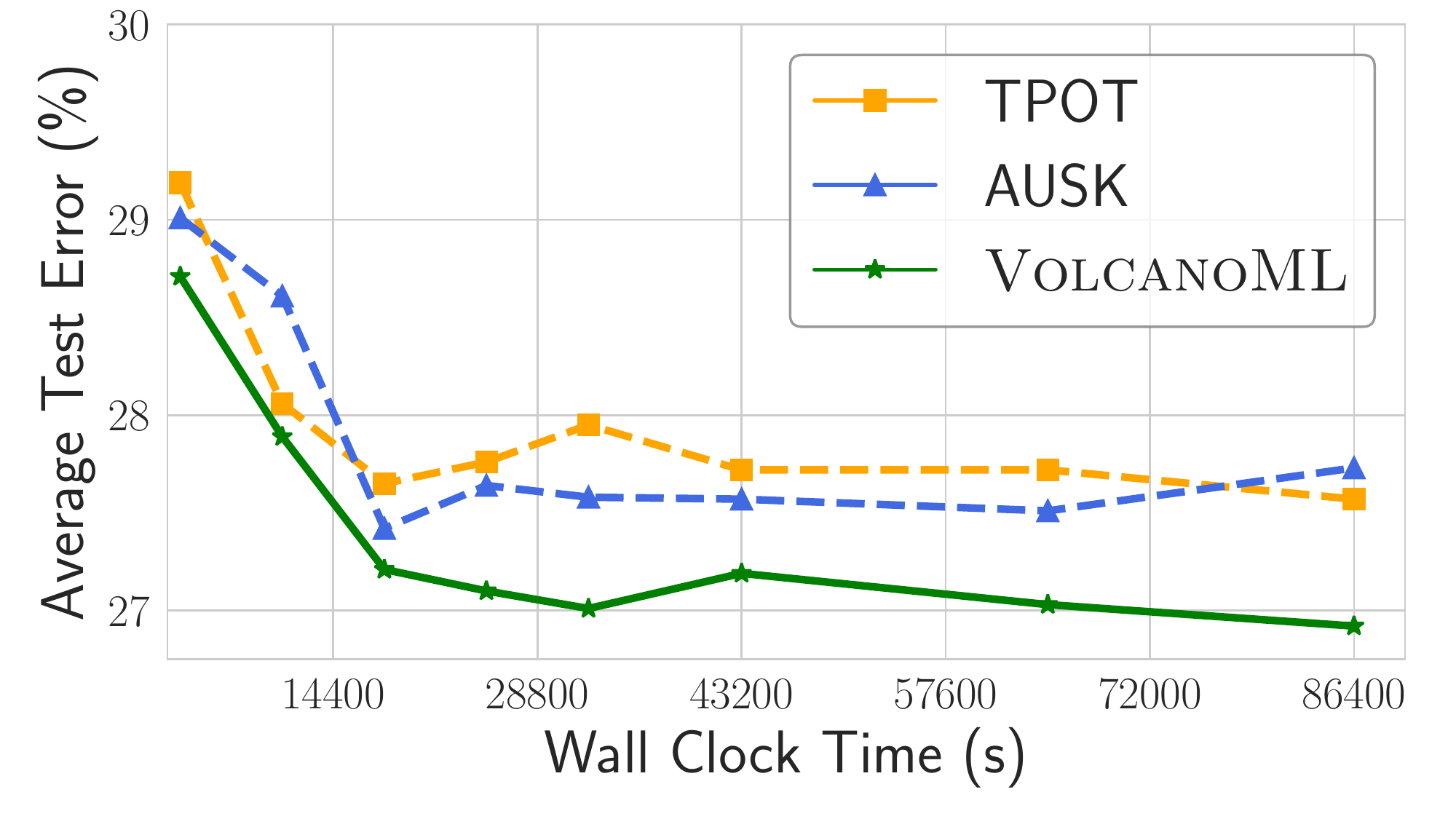}
	}}
	\vspace{-2em}
	\caption{Average test errors on four large datasets with different time budgets.}
	\vspace{-1.5em}
  \label{fig:main_test_speedups}
\end{figure*}

\begin{figure}
	\centering
	\subfigure[Influence Network]{
		\scalebox{0.31}{
			\includegraphics[width=1\linewidth]{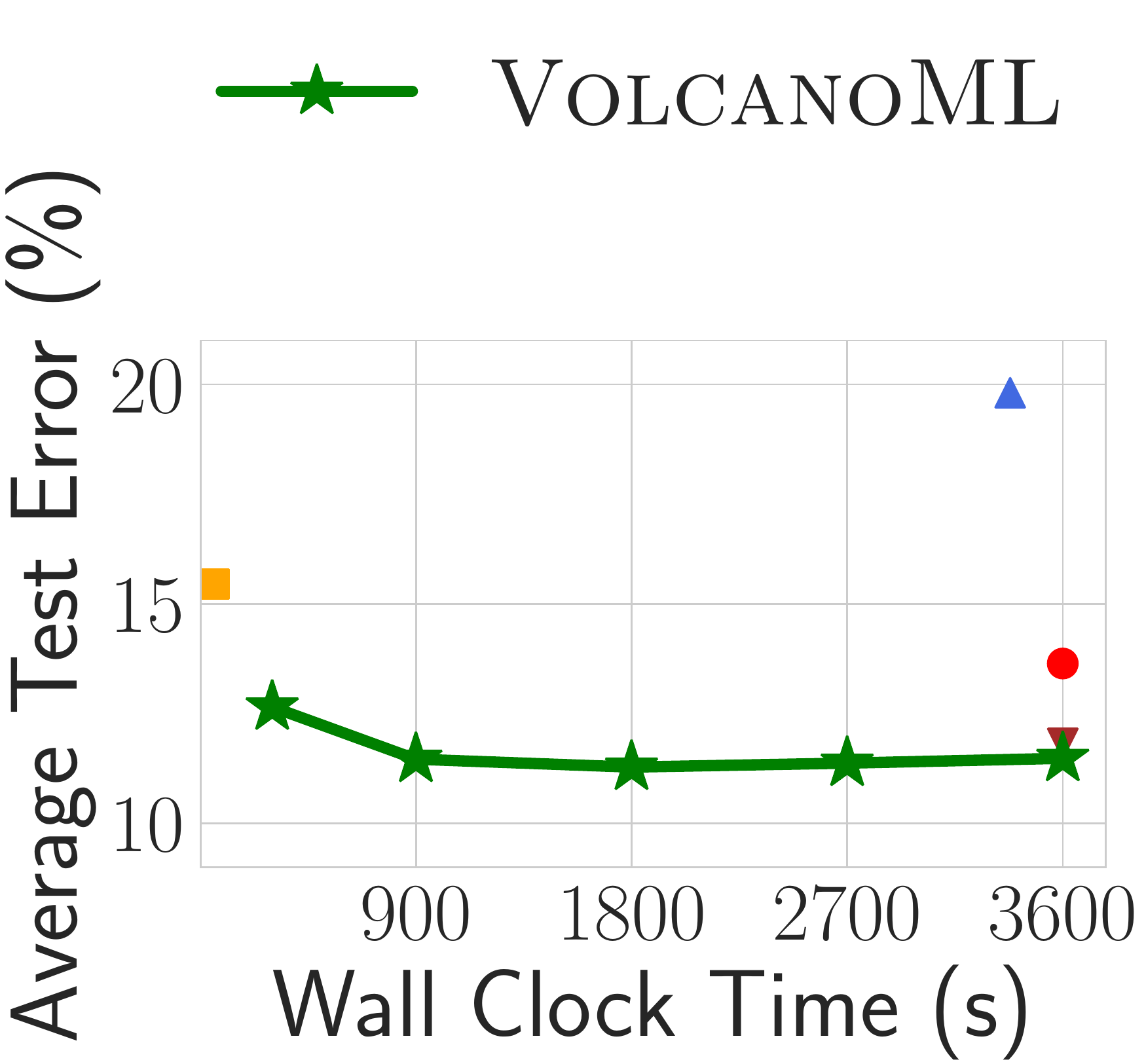}
	}}
	\subfigure[Virus Prediction]{
		\scalebox{0.31}{
			\includegraphics[width=1\linewidth]{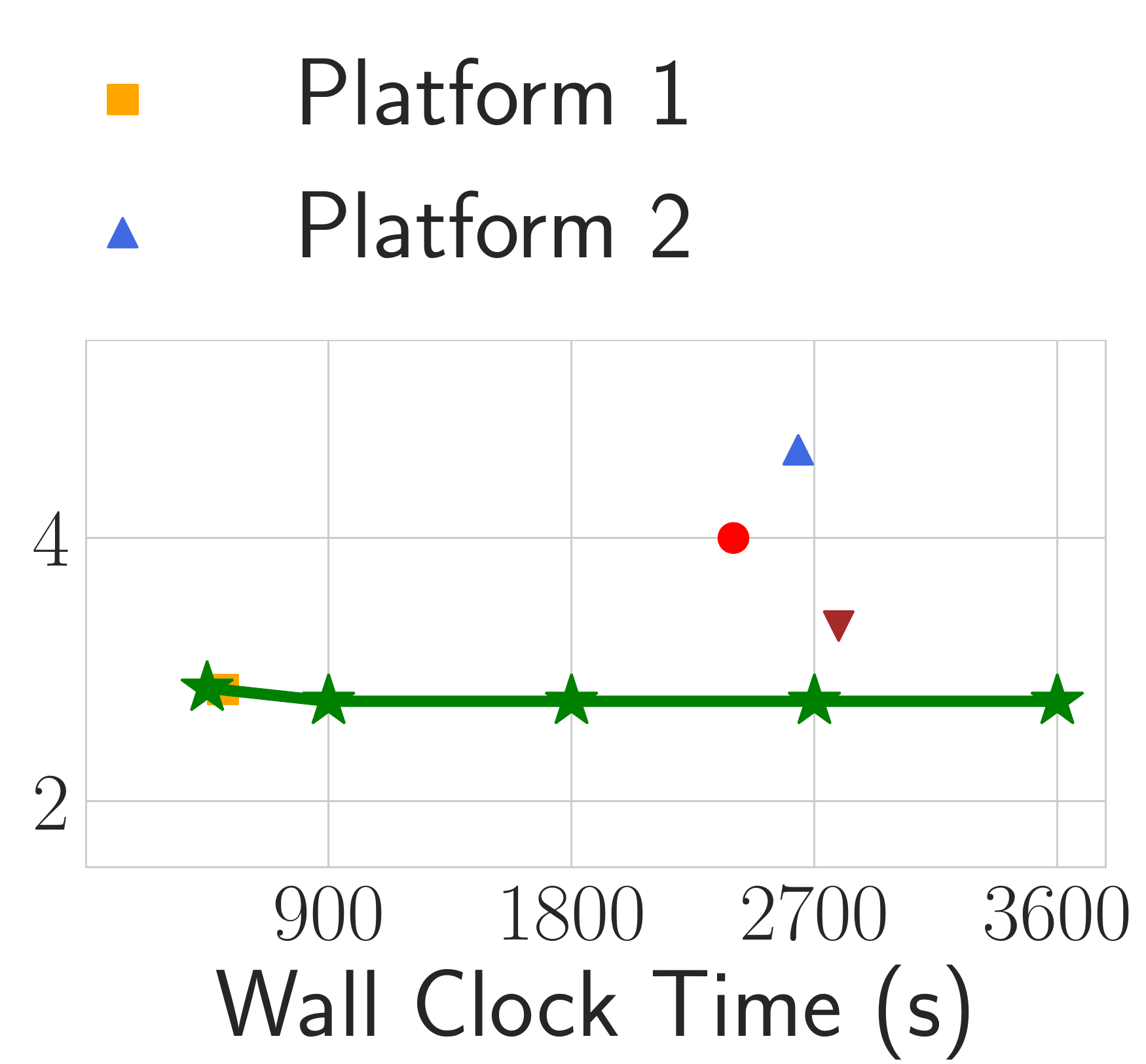}
	}}
	\subfigure[Employee Access]{
		\scalebox{0.31}{
			\includegraphics[width=1\linewidth]{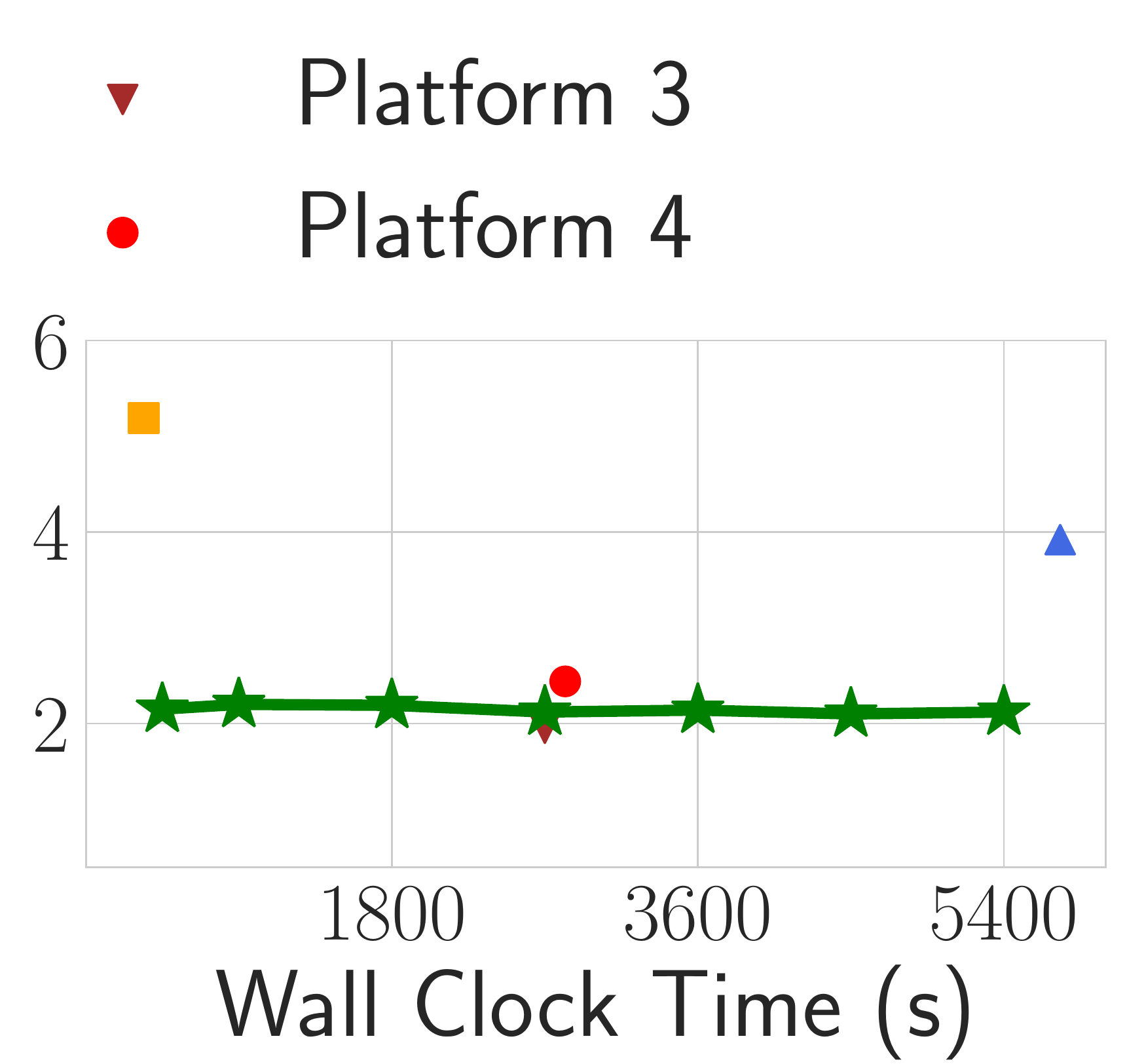}
	}}
    \subfigure[Customer Satisfaction]{
		\scalebox{0.31}{
			\includegraphics[width=1\linewidth]{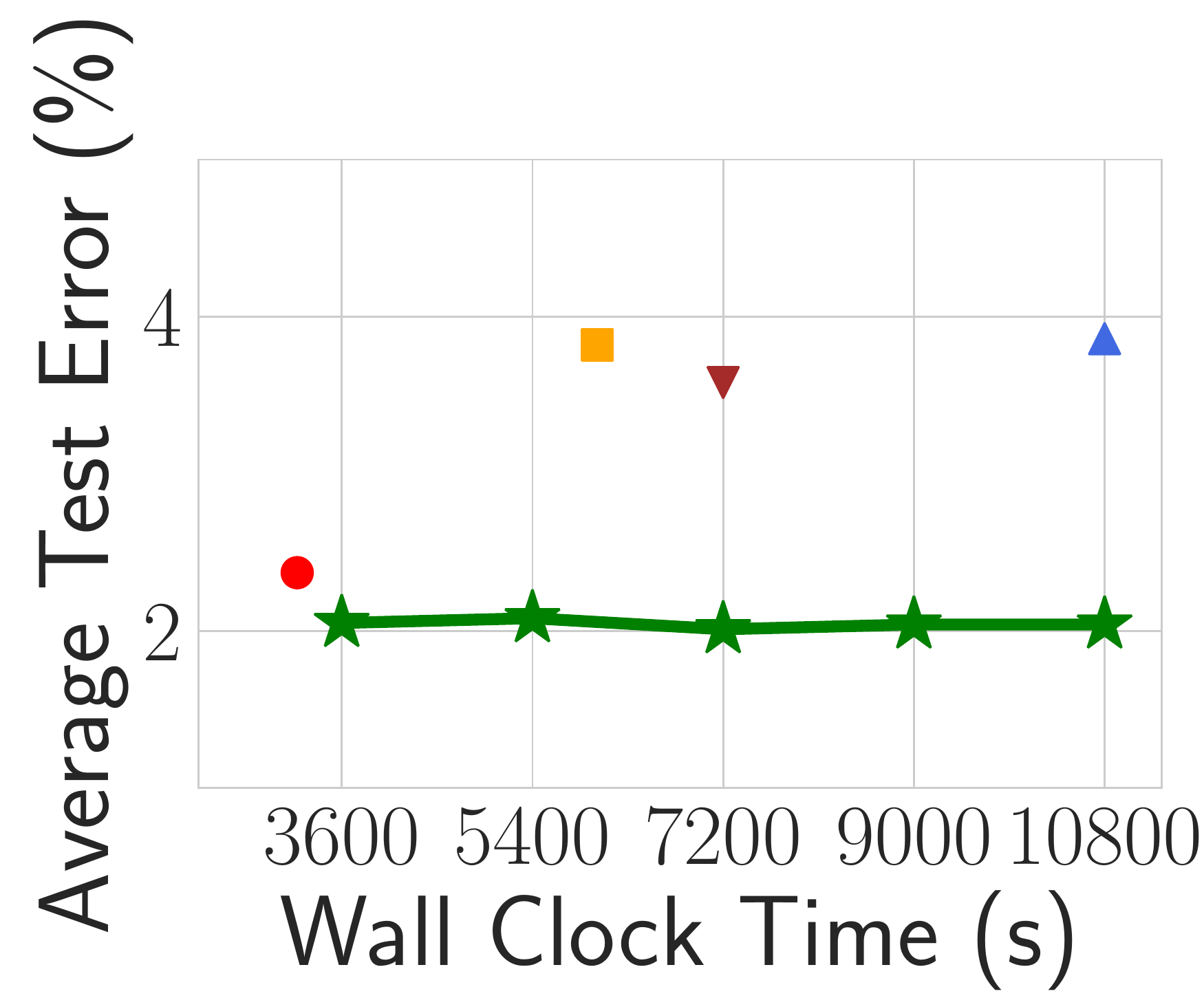}
	}}
	\subfigure[Business Value]{
		\scalebox{0.31}{
			\includegraphics[width=1\linewidth]{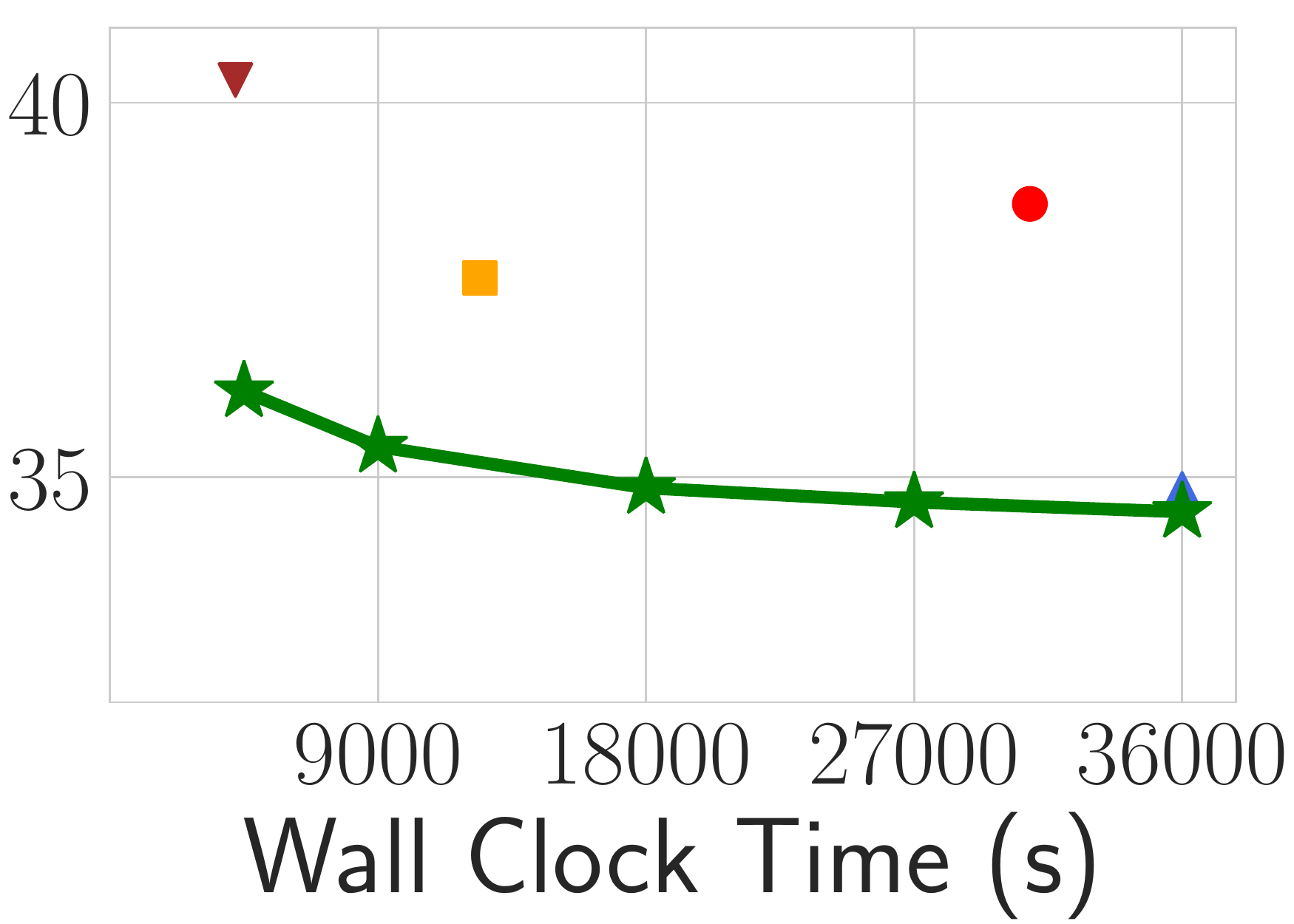}
	}}
	\subfigure[Flavours]{
		\scalebox{0.31}{
			\includegraphics[width=1\linewidth]{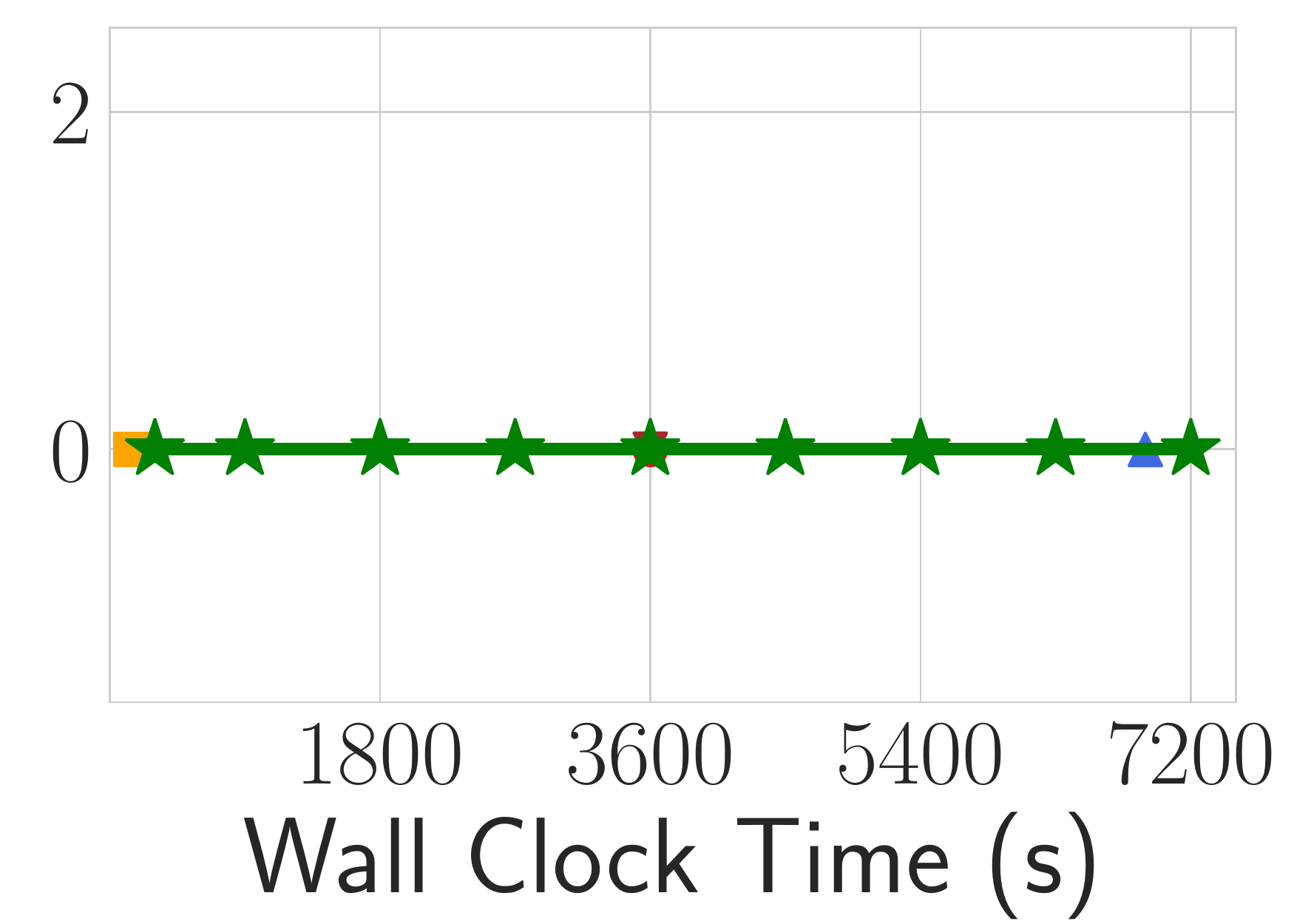}
	}}
	\vspace{-1.5em}
	\caption{Test errors on 6 Kaggle competitions compared with four commercial platforms.}
  \label{fig:industrial_results}
\end{figure}

\vspace{-0.5em}
\subsection{Experimental Setup}
\para{\textbf{AutoML Systems.}}
We evaluate \sys as well as two open source AutoML systems: \texttt{auto-sklearn}~\cite{feurer2015efficient} and \texttt{TPOT}~\cite{olson2019tpot}.
In addition, we also compare \sys with four commercial AutoML platforms from Google, Amazon AWS, Microsoft Azure, and Oracle. 
Both \sys and \texttt{auto-sklearn} support meta-learning, while \texttt{TPOT} does not.
For fair comparison with \texttt{TPOT}, we also use $\sys^{-}$ and AUSK$^{-}$ to denote the versions of \sys and \texttt{auto-sklearn} when meta-learning is disabled.
Our implementation of \sys is available at \url{https://github.com/VolcanoML}.

\para{\textbf{Datasets.}}
To compare \sys with academic baselines, we use 60 real-world ML datasets from the OpenML repository~\cite{10.1145/2641190.2641198}, including 40 for classification (CLS) tasks and 20 for regression (REG) tasks. 
10 of the 40 classification datasets are relatively large, each with 20k to 110k data samples; the other 30 are of medium size, each with 1k to 12k samples. 
In addition, we also use datasets from six Kaggle competitions 
to compare \sys with four commercial platforms.

\para{\textbf{AutoML Tasks.}}
We define three kinds of real-world AutoML tasks, including (1) a general classification task on 30 medium datasets, (2) a general regression task on 20 medium datasets, and (3) a large-scale classification task on 10 large datasets. 

To test the scalability of the participating systems, we design three search spaces that include 20, 29, and 100 hyper-parameters, where the smaller search space is a subset of the larger one.
We run \sys and the baseline AutoML systems against each of the three search spaces.
The time budget is 900 seconds for the smallest search space and 1,800 seconds for the other two, when performing the general classification task (1); the time budget is increased to 5,400 and 86,400 seconds respectively, when performing the general regression task (2) and the large-scale classification task (3).


\para{\textbf{Utility Metrics.}}
Following~\cite{feurer2015efficient}, we adopt the metric \emph{balanced accuracy} for all classification tasks --- compared with standard (classification) accuracy, it assigns equal weights to classes and takes the average of class-wise accuracy.
For regression tasks, we use the \emph{mean squared error} (MSE) as the metric.

In our evaluation, we repeat each experiment 10 times and report the average utility metric.
In each experiment, we use four fifths of the data samples in each dataset to search for the best ML pipeline and report the utility metric on the remaining fifth.

\begin{table}
\scriptsize
\centering
\begin{tabular}{l|ccccc}
\toprule
Search Space - Task & TPOT & AUSK$^{-}$ & AUSK & $\sys^{-}$  & \sys \\
\hline
Small - CLS   & 3.09  & 3.07 & 3.01 & 2.94 & \textbf{2.89}  \\
Medium - CLS  & 3.2  & 3.32  & 3.27  & 2.78  & \textbf{2.43}  \\
Large - CLS  & 3.29  & 3.77 & 3.57 & 2.72 & \textbf{1.65} \\
\hline
Small - REG  & \textbf{2.98}  & 3.02 & 3.0 & 3.02 & \textbf{2.98} \\
Medium - REG & 2.95  & 3.3  & 3.12  & \textbf{2.75}  & 2.88  \\
Large - REG  & 3.1   & 3.85 & 3.82 & 2.15 & \textbf{2.08} \\
\bottomrule
  \end{tabular}
  \caption{Average ranks on 30 classification (CLS) datasets and 20 regression (REG) datasets with three different search spaces. (The lower is the better)}
  \label{scalability_result}
\end{table}

\para{\textbf{Methodology for Comparing AutoML Systems.}}
To compare the overall test result of each AutoML system on a wide range of datasets, we use the \emph{average rank} as the metric following~\cite{bardenet2013collaborative}.
For each dataset, we rank all participant systems based on the result of the best ML pipeline they have found so far; we then take the average of their ranks across different datasets.

\para{\textbf{More Details.}}
We include the details of search space and programming API, experiment settings and additional experiments, and reproductions in the Appendix.

\vspace{-0.5em}
\subsection{End-to-End Comparison}
We first evaluate the participant AutoML systems given the search space explored by \texttt{auto-sklearn}. 
Figure~\ref{end2end_cmp} presents the results of \sys compared to \texttt{auto-sklearn} (AUSK) and \texttt{TPOT} on the 30 classification tasks and the 20 regression tasks, respectively.
For classification tasks, we plot the classification accuracy improvement (\%); for regression tasks, we plot the \emph{relative MSE improvement} $\Delta$, which is defined as     $\Delta(m_1, m_2) = \frac{s(m_2) - s(m_1)}{\operatorname{max}(s(m_2), s(m_1))}$, where $s(\cdot)$ is MSE on the test set. 
Overall, \sys outperforms \texttt{auto-sklearn} and \texttt{TPOT} on 25 and 23 of the 30 classification tasks, and on 17 and 15 of the 20 regression tasks, respectively.

We also conduct experiments to evaluate \sys with different \textit{time budgets}. Figure~\ref{fig:main_test_speedups} presents the results on four large classification datasets.
We observe that \sys exhibits consistent performance over different time budgets. 
Notably, on \texttt{Higgs}, \sys achieves $27.2$\% test error within 4 hours, which is better than the performance of the other two systems given 24 hours.

We further study the \textit{scalability} of the participant systems on the three aforementioned search spaces.
Table~\ref{scalability_result} summarizes the results in terms of the average ranks.
We have two observations:
First, without meta-learning, \sys achieves the best average rank for both the classification and regression tasks --- on the small search space (with 20 hyper-parameters), \sys performs slightly better than \texttt{auto-sklearn} and \texttt{TPOT}, and it performs significantly better on the medium (with 29 hyper-parameters) and large (with 100 hyper-parameters) search spaces. 
Second, with meta-learning, the average rank of \sys is dramatically improved compared with \texttt{auto-sklearn}.
Overall, \sys with meta-learning achieves the best result over large search space.
Furthermore, we also design additional experiments to evaluate the consistency of system performance given different (larger) time budgets and search spaces, and more details can be found in Appendix.


\subsection{Search Space Enrichment}
We now focus on evaluating the \emph{extensibility} of \sys via two experiments with \emph{enriched} search spaces.

\textit{Adding \texttt{Data\_Balancing} Operator.}
In the first experiment, we implement ``\texttt{smote\_balancer}'' -- a new feature engineering operator, and incorporate it into the aforementioned \textit{balancing} stage of feature engineering (FE) (Section~\ref{sec:building-blocks:search-space}).
Note that \texttt{auto-sklearn} cannot support this fine-grained enrichment of the search space.
Table~\ref{enrichment_smote} presents the results of \texttt{auto-sklearn}, \sys without enrichment, and \sys with enrichment, on five imbalanced datasets.
We observe that enriching the search space brings further improvement, e.g., \sys with enrichment outperforms \texttt{auto-sklearn} by 3.57\% (balanced accuracy) on the dataset \texttt{pc2}.

\textit{Supporting Embedding Selection.}
In the second experiment, we add a new stage ``embedding selection'' into the FE pipeline, with two candidate embedding-extraction operators (i.e., two pre-trained models).
This allows \sys to deal with images, which are not easily supported by both \texttt{auto-sklearn} and \texttt{TPOT}.
We implement two pre-trained models to generate embeddings for images, and we evaluate \sys with the enriched search space on the Kaggle dataset \texttt{dogs-vs-cats}.
We observe that \sys achieves 96.5\% test accuracy, which is significantly better than 69.7\% obtained by \texttt{auto-sklearn} without considering embeddings.



\begin{table}
\centering
\scriptsize
\begin{tabular}{lccc}
    \toprule
    Dataset & AUSK & $\sys^{-}$ & \sys \\ 
    \midrule
    sick            & 97.29 & 97.31 & \textbf{97.34}\\
    pc2             & 86.70  & 86.91  & \textbf{90.27}\\
    abalone         & 66.86 & 65.97 & \textbf{67.32}\\
    page-blocks(2)  & 94.70 & 95.29 & \textbf{96.69}\\
    hypothyroid(2)  & 99.62 & 99.64 & \textbf{99.64}\\
    \bottomrule
\end{tabular}
\caption{Test accuracy (\%) of \sys with and without the enrichment of ``\texttt{smote\_balancer}'' operator.} 
\vspace{-1.5em}
\label{enrichment_smote}
\end{table}

\subsection{Comparison with 4 Industrial Platforms}
In addition, we run additional experiments on six Kaggle datasets to compare \sys with four commercial AutoML platforms: 1) Google Cloud AutoML, 2) Microsoft Azure Automated ML, 3) Oracle data science, and 4) Amazon AWS Sagemaker AutoPilot.
Here, we anonymously refer to these platforms as Platform 1-4.
Figure~\ref{fig:industrial_results} show the results, and the Appendix contains the experiment details.
We observe that, given the same time budget (i.e., fix the x-axis to some time budget), \sys is at least comparable with, often outperforms, the considered commercial platforms.



\vspace{-0.5em}
\section{Conclusion}
\label{sec:conclusion}

In this paper, we have presented \sys, a scalable and extensible framework that allows users to design decomposition strategies for large AutoML search spaces in an expressive and flexible manner.
\sys introduces novel building blocks akin to relational operators in database systems that enable expressing search space decomposition strategies in a \emph{structured} fashion -- similar to relational execution plans.
Moreover, \sys introduces a Volcano-style execution model, inspired by its classic counterpart that has been widely used for relational query evaluation, to execute the decomposition strategies it yields.
Experimental evaluation demonstrates that \sys can generate more efficient decomposition strategies that also lead to performance-wise better ML pipelines, compared to state-of-the-art AutoML systems.

\clearpage
\balance


\bibliographystyle{ACM-Reference-Format}
\bibliography{reference}


\end{document}